\documentclass{article}

\usepackage{microtype}
\usepackage{graphicx}
\usepackage{subcaption}
\usepackage{booktabs} 
\usepackage{subfiles}

\usepackage{adjustbox}
\usepackage{tabularx}
\usepackage{amsmath}
\usepackage{wrapfig}
\usepackage{pgfplots}
\usepackage{makecell}
\usepackage{caption}
\usepackage{subcaption}
\usepackage{makecell}
\usepackage{colortbl}

\usepackage{multirow}

\usepackage{hyperref}

\usepackage[preprint]{icml2026}

\usepackage{amsmath}
\usepackage{amssymb}
\usepackage{mathtools}
\usepackage{amsthm}
\usepackage{xspace}

\usepackage[capitalize,noabbrev]{cleveref}

\theoremstyle{plain}

\theoremstyle{definition}

\theoremstyle{remark}

\usepackage[textsize=tiny]{todonotes}

\icmltitlerunning{SHED Light on Segmentation for Dense Prediction}

\begin{document}

\twocolumn[
  \icmltitle{SHED Light on Segmentation for Dense Prediction}



  \icmlsetsymbol{equal}{*}

  \begin{icmlauthorlist}
    \icmlauthor{Seung Hyun Lee}{umich}
    \icmlauthor{Sangwoo Mo}{postech}
    \icmlauthor{Stella X. Yu}{umich}
  \end{icmlauthorlist}

  \icmlaffiliation{umich}{University of Michigan}
  \icmlaffiliation{postech}{POSTECH}

  \icmlcorrespondingauthor{Stella X. Yu}{stellayu@umich.edu}

  \icmlkeywords{Machine Learning, ICML}

  \vskip 0.3in
]



\printAffiliationsAndNotice{}  

\newcommand{\sname}{SHED\xspace}

\definecolor{verylightgray}{gray}{0.92}

\begin{abstract}
Dense prediction infers per-pixel values from a single image and is fundamental to 3D perception and robotics. Although real-world scenes exhibit strong structure, existing methods treat it as an independent pixel-wise prediction, often resulting in structural inconsistencies. We propose SHED, a novel encoder-decoder architecture that enforces geometric prior explicitly by incorporating segmentation into dense prediction. By bidirectional hierarchical reasoning, segment tokens are hierarchically pooled in the encoder and unpooled in the decoder to reverse the hierarchy. The model is supervised only at the final output, allowing the segment hierarchy to emerge without explicit segmentation supervision. SHED improves depth boundary sharpness and segment coherence, while demonstrating strong cross-domain generalization from synthetic to the real-world environments. Its hierarchy-aware decoder better captures global 3D scene layouts, leading to improved semantic segmentation performance. Moreover, SHED enhances 3D reconstruction quality and reveals interpretable part-level structures that are often missed by conventional pixel-wise methods.
\end{abstract}

\vspace{-15pt}

\section{Introduction}

\begin{figure*}[t]
  \centering
  \includegraphics[width=0.9\textwidth]{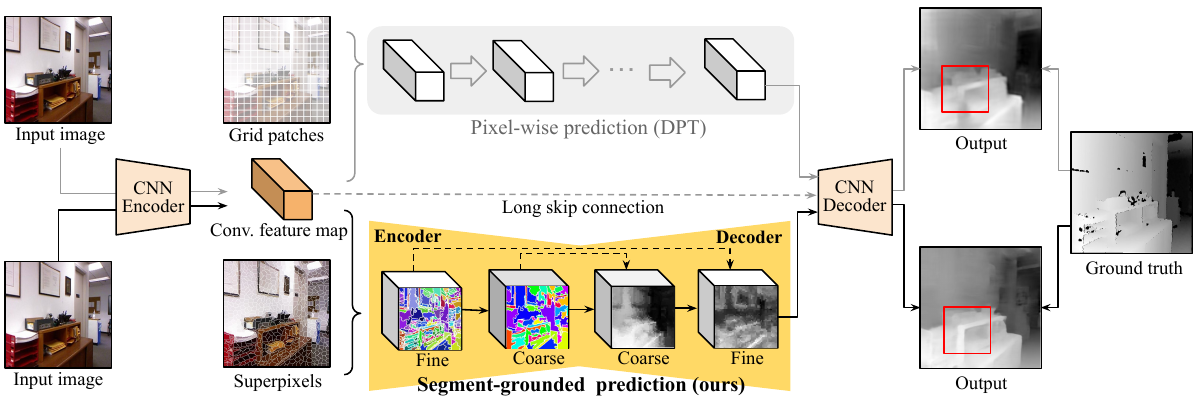}
  \caption{
\textbf{Segment hierarchy for dense prediction (SHED).}
Conventional methods such as DPT~\citep{ranftl2021vision} perform pixel-wise prediction without considering structure, often resulting in blurry object shapes. \sname addresses this by leveraging a hierarchy of segment tokens to guide prediction. Unlike DPT, which uses fixed grid tokens across all layers, we adapt its ViT~\citep{dosovitskiy2020image} blocks into two stages: the encoder pools superpixel tokens into coarser segment tokens, and the decoder progressively refines predictions from coarse to fine segments, producing depth maps with structural coherence. 
  }\label{fig:compare}
  \vspace{-10pt}
\end{figure*}

Dense prediction tasks, such as monocular depth estimation and semantic segmentation, serve as the foundation for spatial reasoning in robotics~\cite{liu2024visual}. At their core, these tasks require an understanding of how 2D image projections relate to the underlying 3D structure of the world. However, modern state-of-the-art models based on Vision Transformers (ViTs)~\citep{dosovitskiy2020image} and pixel-wise regression often treat these tasks as a collection of independent point estimations rather than a coherent structural reconstruction. This disconnect frequently results in structural leakage, where predicted boundaries do not align with physical objects, leading to blurred edges and inconsistent surface geometry~(\cref{fig:compare}, row 1).

In contrast, the human visual system  does not process images as an array of independent points. Instead, it employs a sophisticated bidirectional hierarchy~\citep{hochstein2002view}. Our perception integrates depth and segmentation through a feedback loop where a global scene layout is first inferred by grouping fine-grained textures into parts and wholes. This high-level structural scaffold then guides the refinement of local details. This part-whole reasoning ensures that local pixel-level predictions are constrained by the global geometric context. This organization results in depth maps characterized by sharp, high-fidelity boundaries and smooth intra-object variations (\cref{fig:compare}, row 2).

To realize this idea, we propose \textbf{\sname}, \textbf{S}egment \textbf{H}i\textbf{e}rarchy for \textbf{D}ense Prediction, a novel architecture that re-conceptualizes dense prediction as a process of hierarchical structural reasoning. Unlike traditional models that decode latent features directly back to a rigid pixel grid, \sname learns an implicit hierarchy of segments that adaptively decompose the scene. Note that these segments are not provided via external supervision. They are learned end-to-end, guided solely by the dense prediction objective.

The technical core of \sname lies in its symmetric bidirectional flow. While the encoder iteratively groups pixels into increasingly abstract segment tokens, our contribution focuses on the structural inversion of this process in the decoder. To produce a structured dense map, \sname's decoder unpools features from coarser segments to finer ones using probabilistic assignments computed during the encoding phase. This reverse hierarchy, which has no counterpart in encoder-only segmentation frameworks, enables coarse structural context to directly constrain fine-grained predictions. By distributing segment-level features across associated regions, the model naturally enforces sharp boundaries between distinct objects and maintains geometric smoothness within a single surface. This mechanism allows high-level geometric priors to strictly inform the pixel-level output, preserving the global layout while capturing fine-grained structural detail.

Integrating hierarchical segmentation into the dense prediction pipeline grants \sname three key advantages: 
\vspace{-1em}
\begin{itemize}
    \item \sname produces depth maps with high-fidelity boundary alignment. This inherent structural bias significantly enhances robustness in cross-domain transfer, such as from synthetic to real-world environments, where traditional pixel-wise models often fail to maintain consistency.
    \vspace{-0.5em}
    \item We demonstrate that depth supervision induces structured representations that capture the 3D scene layout. Consequently, \sname refines semantic segmentation by ensuring that category labels strictly adhere to geometrically consistent spatial boundaries.
    \vspace{-0.5em}
    \item Since \sname maintains a symbolic hierarchy of segments, it enables the discovery of object parts in 3D space without any part-level labels. This provides a level of structural interpretability that holistic, black-box architectures like DPT \citep{ranftl2021vision} fundamentally lack.
    \vspace{-0.8em}
\end{itemize} 


\section{Related Work}



\textbf{Dense prediction} is a core problem in computer vision, aiming to assign pixel-level outputs across an image~\citep{forsyth2002computer}. It includes tasks such as semantic segmentation~\citep{long2015fully}, depth estimation~\citep{eigen2014depth}, optical flow~\citep{teed2020raft}, and image editing~\citep{isola2017image}. Modern approaches typically adopt encoder-decoder architectures, such as U-Net~\citep{ronneberger2015u} and DPT~\citep{ranftl2021vision}, trained using task-specific supervision. These models perform well on benchmarks focused on per-pixel accuracy, as demonstrated by large-scale systems like Segment Anything~\citep{ravi2025sam} and Depth Anything~\citep{yang2024depth2}. However, recent studies show that even top-performing models often lack structural consistency~\citep{el2024probing, man2024lexicon3d}. We argue that dense prediction should move beyond local estimation toward structured reasoning guided by region-level abstraction.


\begin{figure*}[t]
  \centering
  \includegraphics[width=0.95\linewidth]{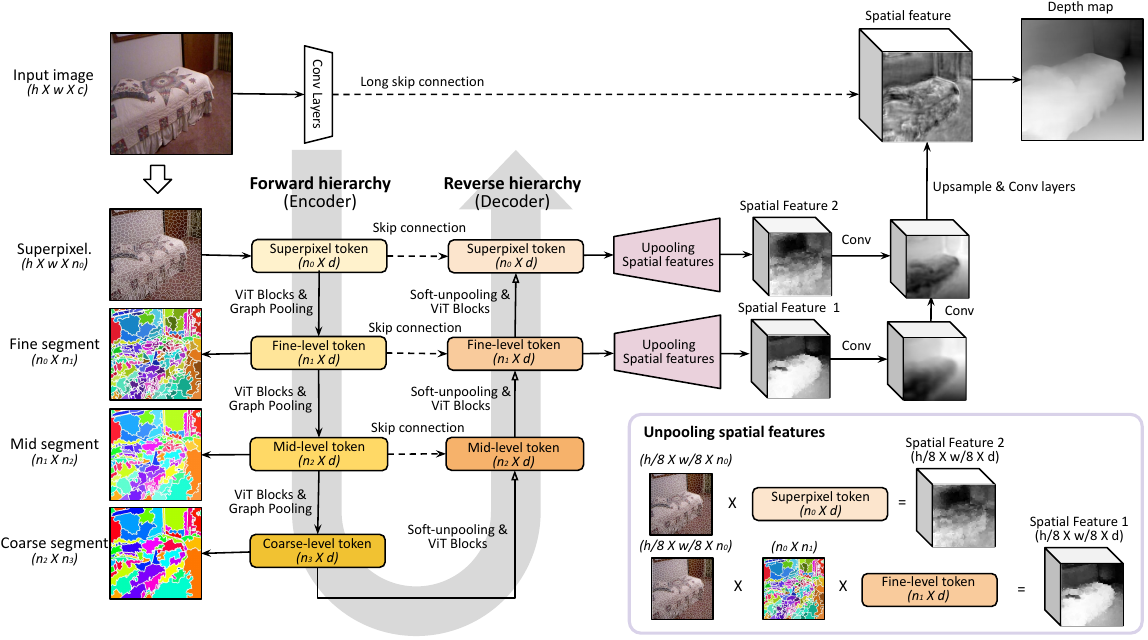}
\caption{\textbf{\sname introduces a bidirectional segment hierarchy for dense prediction.}
\sname decomposes dense prediction into two complementary processes: structural abstraction and structural inversion.
\textbf{1)} The encoder constructs a forward hierarchy by grouping superpixel tokens into increasingly abstract segments.
\textbf{2)} The decoder explicitly inverts this hierarchy by unpooling segment representations from coarse to fine, enabling global structural context to directly constrain pixel-level predictions.
Segment tokens are projected into region-aligned 2D feature maps and fused across hierarchy levels, together with early convolutional features, to recover fine details and produce the final dense map.
}
\vspace{-10pt}
\label{fig:method}
\end{figure*}

\textbf{Monocular depth estimation} is a representative dense prediction task, that infers per-pixel depth from a single image. It is widely used in 3D reconstruction~\citep{song2017semantic}, autonomous driving~\citep{geiger2012we}, and robotic perception~\citep{tateno2017cnn}. 
Structural cues in depth estimation have been extensively explored to enhance geometric coherence. Existing approaches can be broadly categorized into four types:
\textbf{1)} Representation approaches modify how depth is encoded, such as by discretizing depth values~\citep{fu2018deep,bhat2021adabins,li2022binsformer} or modeling spatial dependencies \citep{liu2015learning,cheng2018depth,yuan2022neural}. 
\textbf{2)} Regularization imposes geometric constraints through loss functions that promote smooth surfaces~\citep{godard2017unsupervised,zhan2018unsupervised,bian2019unsupervised}, consistent normals~\citep{yang2018unsupervised}, or planar regions~\citep{yin2019enforcing,watson2019self}. 
\textbf{3)} Multi-task learning jointly estimates depth with auxiliary signals, such as scene geometry~\citep{eigen2015predicting,yin2018geonet} or semantics~\citep{mousavian2016joint,kendall2018multi,chen2019towards,guizilini2020semantically,zhu2020edge}.  
\textbf{4)} Post-processing refines predictions using off-the-shelf techniques~\citep{krahenbuhl2011efficient,chen2016single}.

Several multi-task approaches have explored segmentation as an auxiliary signal to improve depth estimation. Early works used segmentation as an additional supervision signal~\citep{mousavian2016joint,kendall2018multi}, while more recent ones leveraged segment regions or boundaries to guide depth discontinuities~\citep{chen2019towards,guizilini2020semantically,zhu2020edge}.
In contrast, \sname integrates segmentation directly into the model’s internal representation, enabling depth and segmentation to be jointly constructed and refined within a unified hierarchical framework. Moreover, the segment hierarchy in \sname is learned in an unsupervised manner, eliminating the need for additional human annotations.

\textbf{Perceptual grouping} is a key mechanism in human vision that organizes low-level elements into coherent global structures~\citep{wertheimer1938laws,marr2010vision}. This principle has inspired a broad range of computer vision research, including perception~\citep{locatello2020object,mo2021object,kang2022oamixer,deng2023perceptual,ranasinghe2023perceptual}, segmentation~\citep{arbelaez2012semantic,hwang2019segsort,ke2022unsupervised,xu2022groupvit}, and generation~\citep{hong2018inferring,mo2018instagan,he2022ganseg}. In particular, CAST~\citep{ke2024learning} recently applied it to ViTs for concurrent segmentation and recognition. However, most of these methods, including CAST, consider only a \textit{forward hierarchy}, constructing representations and segmentations in a bottom-up manner. Without a reverse hierarchy, such representations cannot enforce spatial consistency in fine-grained dense prediction. In contrast, we adopt the complementary concept of a \textit{reverse hierarchy}~\citep{hochstein2002view}, where global structures guide and refine local parts through top-down feedback. While some prior works~\citep{anderson2018bottom,shi2023top,eftekhar2023selective} have explored reverse hierarchies for recognition, they do not address dense prediction. Other studies~\citep{eslami2016attend,sajjadi2022object,seitzer2022bridging} apply similar ideas to encoder-decoder architectures, but focus on object-centric representations, lacking the ability to model segment hierarchies. To the best of our knowledge, this is the first work to leverage bidirectional segment hierarchies to enhance dense prediction within a modern ViT.


\section{Segment Hierarchy for Dense Prediction}

\label{sec:method}



We propose \sname, a dense prediction architecture that explicitly models a bidirectional segment hierarchy (see \cref{fig:method}).
The encoder groups pixels into segment tokens and progressively pools them into coarser representations, while the decoder reverses this process by unpooling segment features from coarse to fine.
By propagating segment-level structure back to pixel-level predictions, \sname enables global scene layout to directly constrain dense outputs, overcoming a limitation of pixel-wise models.


\vspace{-5pt}

\subsection{Encoder: Grouping segments via forward hierarchy}
\vspace{-5pt}

The encoder constructs a forward segment hierarchy that abstracts fine-grained regions into increasingly coarse structural units.
This hierarchy serves as the structural backbone of \sname, capturing part–whole relationships that are difficult to represent using grid-based tokens.
While similar hierarchical grouping strategies have been explored for image-level recognition, dense prediction requires such abstractions to be explicitly reversible, which motivates our overall bidirectional design.

To construct the forward hierarchy, we adopt a segment-based tokenization strategy that replaces square patch tokens with superpixel tokens and progressively clusters them based on feature similarity.
This grouping mechanism follows prior work on hierarchical segmentation~\cite{ke2024learning}, but in \sname it serves a fundamentally different role: providing intermediate structural representations that will later be inverted for dense prediction.

Unlike encoder-only hierarchical segmentation models, the forward hierarchy in \sname is not an end in itself.
Instead, it defines an intermediate structural representation whose primary purpose is to be inverted by the decoder to enable structured dense prediction.


\begin{figure*}[t]
  \centering\footnotesize
  \setlength{\tabcolsep}{1pt}   
  \renewcommand{\arraystretch}{1.0}

  \newcommand{\hspacesmall}{\hspace{2pt}}
  \newcommand{\hspacelarge}{\hspace{2pt}}


\begin{tabular}{
   @{}
   c@{\hspacesmall}c@{\hspacelarge}
   c@{\hspacesmall}c@{\hspacesmall}c@{\hspacelarge}
   c@{\hspacesmall}c@{}
   }
    Image & Superpixel & \multicolumn{3}{c}{Segments (fine-to-coarse)} & Prediction & GT \\[1pt]

    \includegraphics[width=0.138\textwidth]{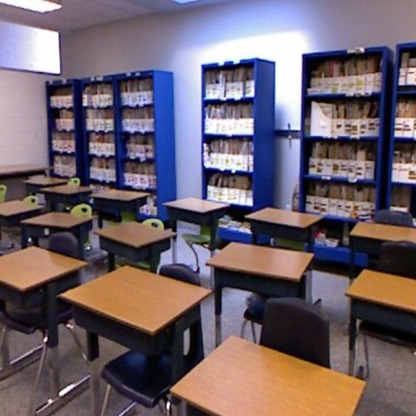} &
    \includegraphics[width=0.138\textwidth]{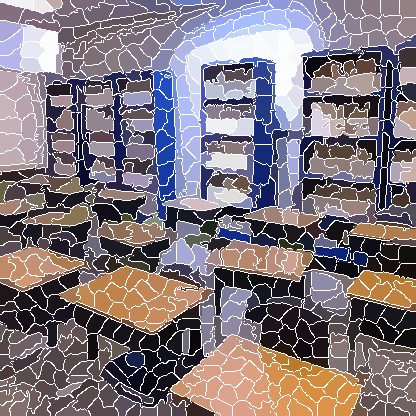} &
    \includegraphics[width=0.138\textwidth]{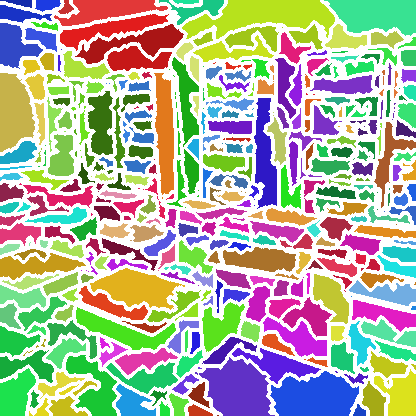} &
    \includegraphics[width=0.138\textwidth]{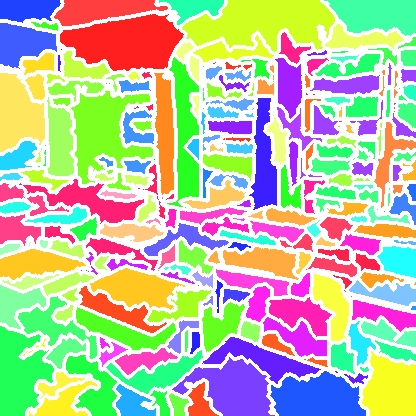} &
    \includegraphics[width=0.138\textwidth]{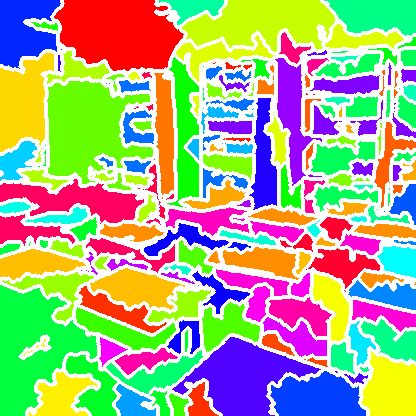} &
    \includegraphics[width=0.138\textwidth]{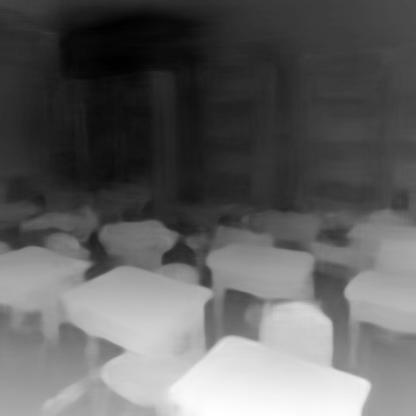} & 
    \includegraphics[width=0.138\textwidth]{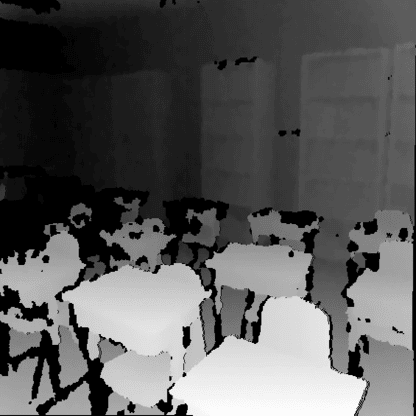} \\

    \includegraphics[width=0.138\textwidth]{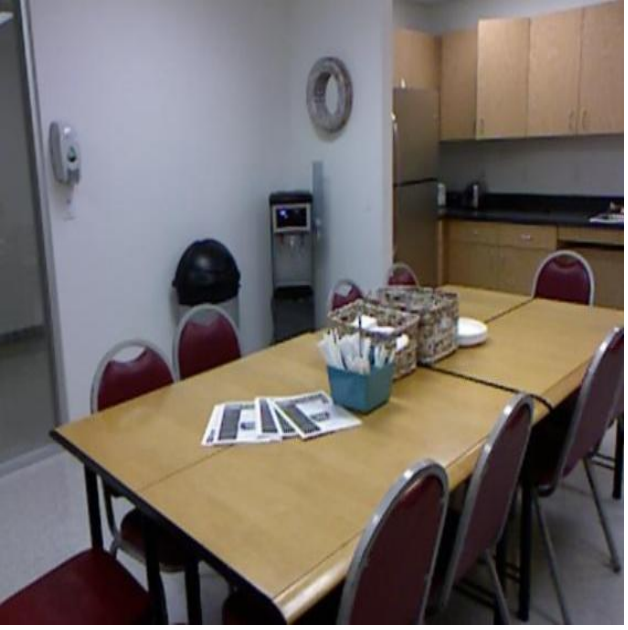} &
    \includegraphics[width=0.138\textwidth]{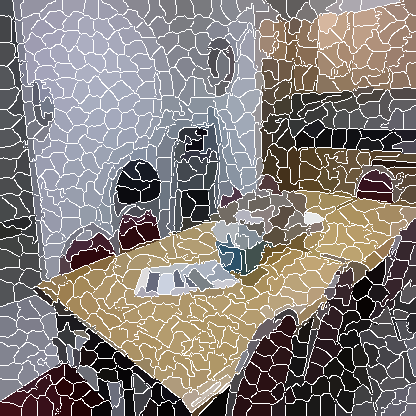} &
    \includegraphics[width=0.138\textwidth]{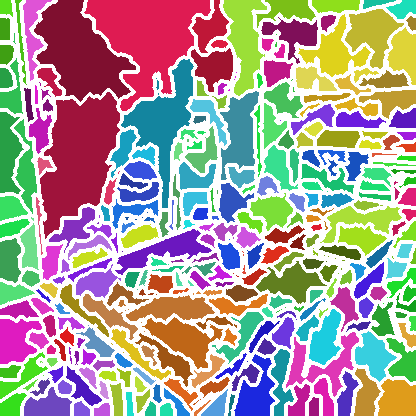} &
    \includegraphics[width=0.138\textwidth]{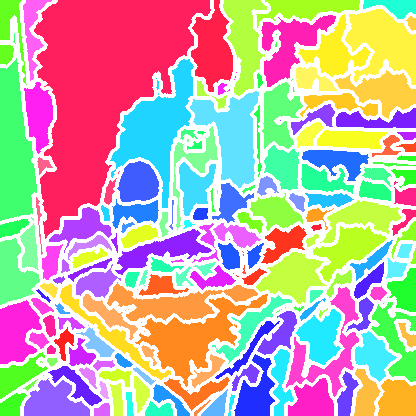} &
    \includegraphics[width=0.138\textwidth]{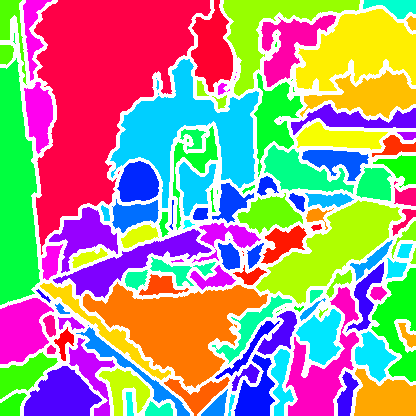} &
    \includegraphics[width=0.138\textwidth]{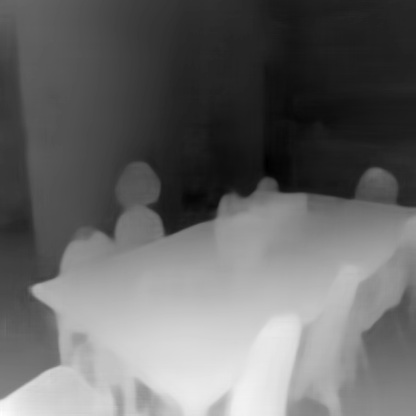} & 
    \includegraphics[width=0.138\textwidth]{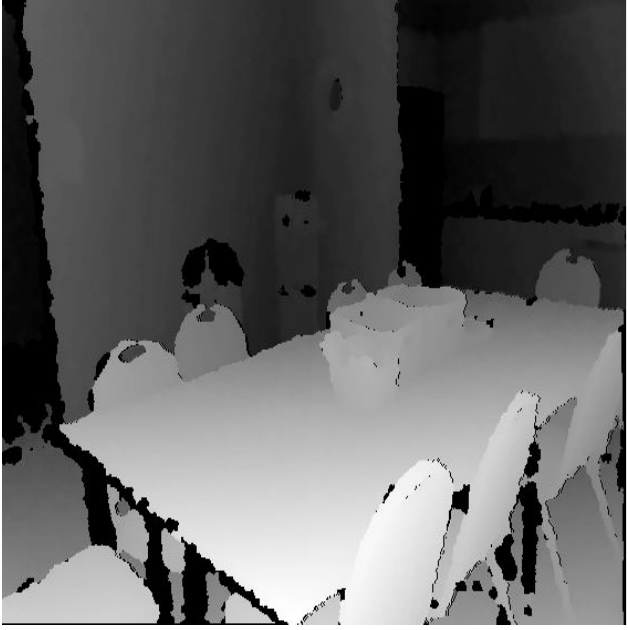} \\ 
  
    \end{tabular} 

\vspace{-5pt}
\caption{
\textbf{\sname produces consistent structures in predicted depth map with spatio-layout.}  
We visualize the fine-to-coarse segments and corresponding depth maps from \sname, along with ground truth (GT) depth. Examples are from the NYUv2 test set.  
\sname captures fine structures through its segments, such as desks in a classroom, which allow the depth map to clearly separate them from the background (row 1). It also decomposes large objects, such as a table, into multiple parts, leading to smooth depth variations toward the back (row 2). 
}\label{fig-main}
\vspace{-5pt}
\end{figure*}

\begin{figure*}[t]
  \centering\footnotesize
  \setlength{\tabcolsep}{1pt}   
  \renewcommand{\arraystretch}{1.0}

  \newcommand{\hspacesmall}{\hspace{2pt}}
  \newcommand{\hspacelarge}{\hspace{2pt}}
 
  \begin{tabular}{
  @{}
  c@{\hspacelarge}
  c@{\hspacesmall}c@{\hspacesmall}c@{\hspacelarge}
  c@{\hspacesmall}c@{\hspacesmall}c@{}
  }
    Image &
    DPT &
    \sname (ours) &
    GT &
    DPT & 
    \sname (ours) 
    \\
    \includegraphics[width=0.16\textwidth]{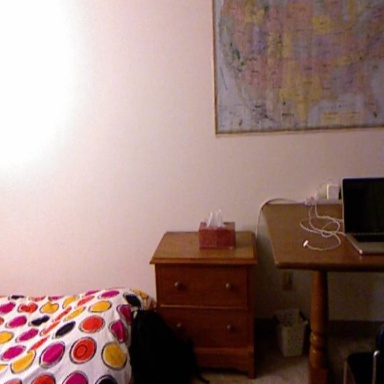} &
    \includegraphics[width=0.16\textwidth]{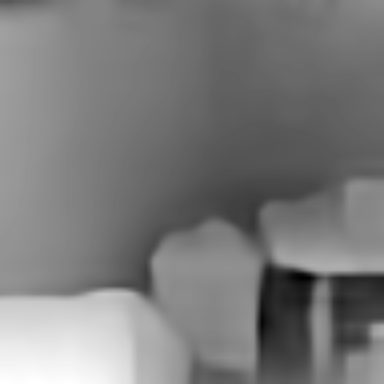} &
    \includegraphics[width=0.16\textwidth]{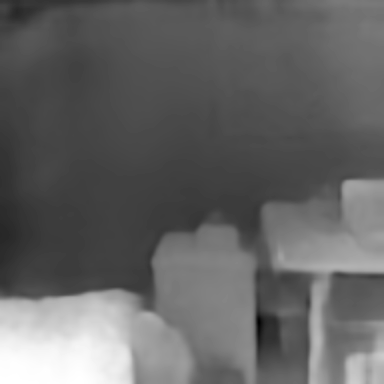} &
    \includegraphics[width=0.16\textwidth]{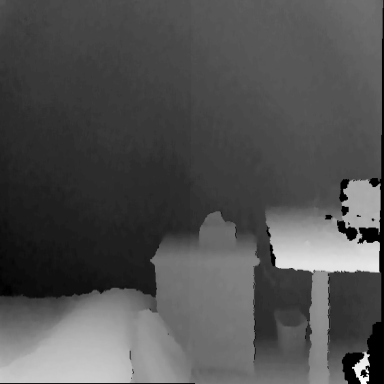} &
    \includegraphics[width=0.16\textwidth]{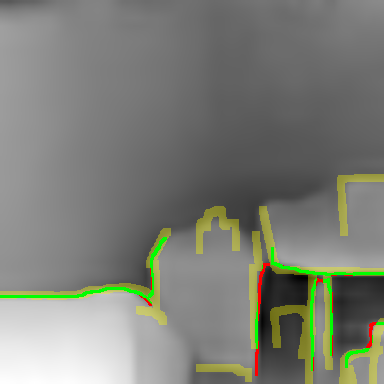} &
    \includegraphics[width=0.16\textwidth]{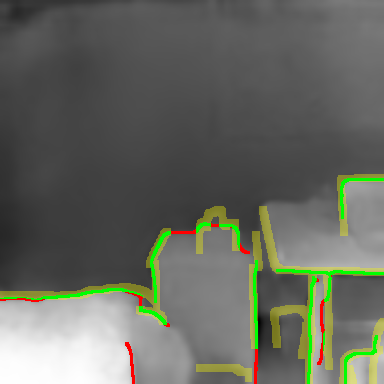} & 
    \\
    \includegraphics[width=0.16\textwidth]{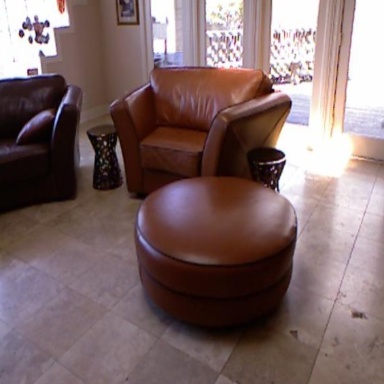} &
    \includegraphics[width=0.16\textwidth]{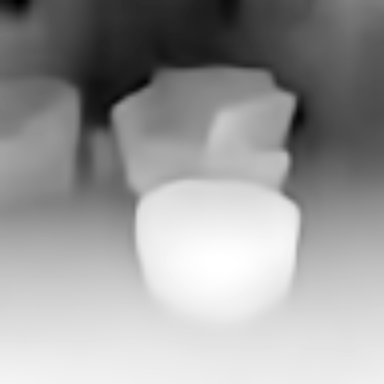} &
    \includegraphics[width=0.16\textwidth]{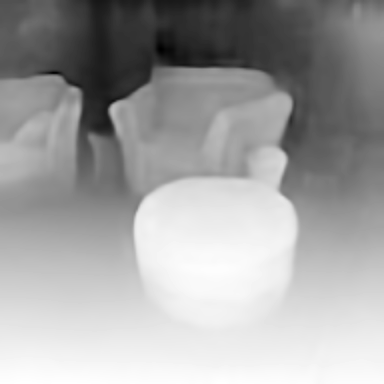} &
    \includegraphics[width=0.16\textwidth]{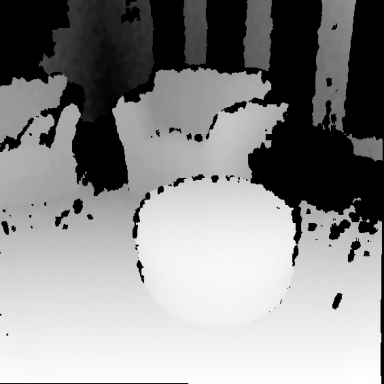} &
    \includegraphics[width=0.16\textwidth]{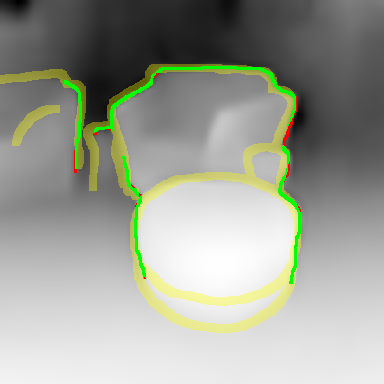} &
    \includegraphics[width=0.16\textwidth]{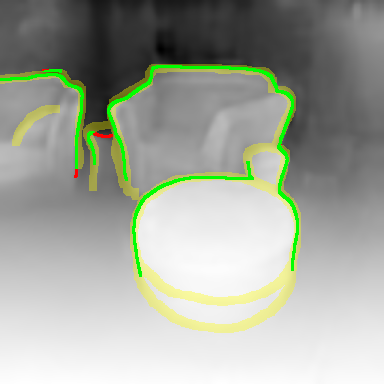} & 
  \end{tabular}

\vspace{-5pt}
  \caption{
\textbf{\sname generates sharper object contours, clearer occlusion boundaries, and more coherent values within segments.}
We compare depth maps (cols 2-4) and occlusion boundaries (cols 5, 6) from DPT, \sname on the NYUv2-OC++ dataset. Boundaries are extracted using a Canny edge detector and evaluated against GT, with GT edges shown in yellow, true positive in green and false positive in red. \sname more accurately captures object edges and produces smoother depth within segments. Its predicted boundaries also align more closely with the ground truth.
    }\label{fig:boundary}
\vspace{-10pt}
\end{figure*}

\textbf{Tokenization.}
Given an image $X \in \mathbb{R}^{h \times w \times c}$, the encoder produces hierarchical segmentations $S_0, S_1, \dots$ and corresponding embeddings $Z_0, Z_1, \dots$, ordered from fine to coarse. This process begins by dividing the image into $n_0$ superpixels, which yields a one-hot assignment matrix $S_0 \in \mathbb{R}^{(h \cdot w) \times n_0}$ that maps each pixel to a superpixel. We extract a convolutional feature map $F_\text{conv} \in \mathbb{R}^{(h_0 \cdot w_0) \times d}$ with spatial stride 8 ($h_0 = h/8$, $w_0 = w/8$), add fixed sinusoidal positional embeddings, and average-pool features within each superpixel to obtain initial embeddings $Z_0 \in \mathbb{R}^{n_0 \times d}$. For global context modeling, we append a class token to form $\bar{Z}_0 \in \mathbb{R}^{(n_0 + 1) \times d}$, and feed it into the first ViT block.

\textbf{Hierarchical clustering.}
Coarser segments are obtained by alternating ViT blocks and graph pooling~\citep{ke2024learning}. At each level $l$, given $Z_{l-1}$ and $S_{l-1}$ from the previous layer, we append a class token to form $\bar{Z}_{l-1}$, apply ViT blocks, and obtain updated features, excluding the class token.

To form coarser tokens $Z_l \in \mathbb{R}^{n_l \times d}$, we compute a soft assignment matrix $P_l \in \mathbb{R}^{n_{l-1} \times n_l}$ based on cosine similarity between fine- and coarse-level tokens:
\[
P_l(i \rightarrow j) \propto \text{sim}(Z_{l-1}[i], Z_l[j]), \quad \text{for } i \in [n_{l-1}], \, j \in [n_l],
\]
where $[n] := \{0, \dots, n{-}1\}$.
The coarse tokens $Z_l$ are initialized via farthest point sampling~\citep{qi2017pointnet++} from $Z_{l-1}$, and refined by aggregating fine-level features by $P_l$, followed by an MLP and a residual connection:
\[
Z_l \gets Z_l + \text{MLP}(P_l^\top Z_{l-1} \oslash P_l^\top \mathbf{1}),
\]
where $\oslash$ denotes element-wise division for normalization.

To propagate segmentation labels through the hierarchy, we compute coarser segmentations by composing the assignment matrices:
\[
S_l = S_{l-1} \, \bar{P}_l, \quad l = 1, 2, \dots, l_\text{max},
\]
where $\bar{P}_l$ is a hard assignment matrix obtained by taking the argmax over each row of $P_l$.

\vspace{-5pt}
\subsection{Decoder: Predicting outputs via reverse hierarchy}
\vspace{-5pt}

The decoder reconstructs spatial feature maps by reversing the encoder’s segment hierarchy, progressively unpooling segment tokens $Z_{l_\text{max}}, \dots, Z_0$. This involves two steps:  
\textbf{1)} computing decoder features $Z'_l$ by unpooling from $Z'_{l+1}$ and fusing them with encoder features $Z_l$ via skip connections; and  
\textbf{2)} projecting $Z'_l$ to the image space to obtain a spatial feature map $F_l$ of size $(h_l, w_l)$.


\textbf{Unpooling segment tokens.} Formally, the reverse hierarchy distributes coarse segment features to finer segments through learned soft assignments. At each level $l = l_\text{max}-1, \dots, 0$, we compute
\[
Z'_{l} \gets P_{l+1}^\top \, Z'_{l+1},
\]
which distributes coarse features to finer segments. We then add unpooled features with corresponding encoder output:
\[
Z'_l \gets \text{MLP}(Z'_l + Z_l),
\]
followed by ViT blocks with class tokens.

\textbf{Unpooling spatial features.}
We convert the segment tokens $Z'_l$ into spatial feature maps by composing the soft assignment matrices:
\[
P_{0 \rightarrow l} = P_1 \cdots P_l \quad \in \mathbb{R}^{n_0 \times n_l},
\]
and applying them to the initial superpixel-to-pixel map $S_0$ to obtain soft segmentations $S_{0 \rightarrow l} = S_0 P_{0 \rightarrow l}$.
The spatial feature map is then reconstructed as
\[
F_l = S_{0 \rightarrow l} \, Z'_l, \quad F_l \in \mathbb{R}^{(h_l \cdot w_l) \times d}.
\]
The set of spatial maps $\{F_l\}_{l=1}^{l_\text{max}}$ is fused using convolutional layers, combined with $F_\text{conv}$, and refined through final convolution and upsampling to produce the final dense prediction.

A common design in dense prediction is to build a spatial hierarchy by progressively reducing feature resolution.
DPT~\citep{ranftl2021vision} is a representative example of this approach. DPT reduces the spatial resolution of feature maps $F_l$ at each level by a factor of $2^l$, with $h_l = h_0 / 2^l$, $w_l = w_0 / 2^l$, producing coarse maps in early ViT layers that are progressively refined. However, it relies on local aggregation, which lacks structural consistency. In contrast, \sname replaces the spatial hierarchy with a segment hierarchy that explicitly models region abstraction and inversion. By operating on segment tokens rather than grid locations, \sname enforces structural consistency across hierarchy levels and within regions. Therefore, we omit spatial reduction in \sname and simply set $h_l = h_0$, $w_l = w_0$.



\vspace{-5pt}

\section{Experiments}

\vspace{-5pt}

We demonstrate the benefits of \sname by integrating segmentation into the loop for dense prediction: \textbf{1)} Segment-consistent depth estimation that preserves occlusion boundaries and intra-segment coherence, leading to improved accuracy and efficiency; \textbf{2)} Structure-aware representation learning through dense supervision; \textbf{3)} 3D scene reconstruction from predicted depth maps, yielding globally coherent and part-aware structures.

\vspace{-5pt}

\subsection{Setup}

\vspace{-5pt}

For fair comparison with prior dense prediction methods, we follow the standard training protocol commonly used in DPT-style models~\citep{ranftl2021vision}. Specifically, we adopt the DPT-Hybrid configuration, which combines ResNet-50~\citep{he2016deep} and ViT-Small~\citep{dosovitskiy2020image}, and refer to it simply as DPT throughout the paper. For in-domain evaluation, we primarily train and evaluate on NYUv2~\citep{Silberman:ECCV12}, a standard benchmark for indoor depth estimation.  For cross-domain transfer, we train \sname on the synthetic HyperSim dataset~\citep{roberts2021hypersim} and assess its zero-shot generalization on the real-world NYUv2 dataset. We further compare our approach with stronger prior-based models, including the DPT-style Depth Anything v2~\citep{yang2024depth2} and Marigold~\citep{ke2025marigold}, both fine-tuned on HyperSim. Depth Anything v2 uses the DPT decoder with DINOv2~\citep{oquab2023dinov2} encoder.

\textbf{Tokenization.}
Input images of size 640$\times$480 are randomly cropped to 384$\times$384 during preprocessing. We generate 576 superpixels using the SEEDS algorithm~\citep{van2012seeds}, matching the 24$\times$24 token grid of DPT, which corresponds to 16$\times$16 patches. Features are extracted from intermediate ResNet-50 blocks at 1/4 and 1/8 of the input resolution; the latter initializes segment token embeddings, while both are passed to the final decoder via skip connections.  This entire preprocessing and tokenization pipeline is applied consistently in all experiments.

\begin{table*}[t]
\centering\small
\caption{%
\textbf{\sname improves boundary accuracy and object-level consistency.}
We evaluate the structural quality of depth maps using two metrics:
\textbf{1)} Occlusion boundary error, evaluated on the NYUv2-OC++ dataset. Occlusion boundaries are extracted using a Canny edge detector, and the Chamfer distance is computed in both directions: from prediction to ground truth and vice versa. \textbf{2)} Intra-segment coherence measures how well the predicted depth values within each object align with the ground-truth using object-level annotations.
}\label{tab:boundary}
\vspace{-3pt}

\begin{tabular}{@{}l c c c c c c@{}}
\toprule
\multirow{2}{*}{Method} 
& \multicolumn{2}{c}{ Boundary Error $\downarrow$}
& \multicolumn{1}{c}{Object-wise Depth Accuracy $\uparrow$}
& \multicolumn{3}{c}{Object-wise Depth Error $\downarrow$} \\
\cmidrule(lr){2-3}\cmidrule(lr){4-4}\cmidrule(lr){5-7}
& $\epsilon_a$ & $\epsilon_c$ & $\delta > 1.25$ 
& AbsRel & RMSE & $\log {10}$ \\
      \midrule
       DPT & 6.395 & 1.438 &  0.802 & 0.144 & 0.500 & 0.061 \\
    \sname(ours) & \textbf{5.713} & \textbf{0.608} & \textbf{0.814}  & \textbf{0.142} & \textbf{0.496} & \textbf{0.060} \\
      \bottomrule
\end{tabular}
\vspace{7pt}
\caption{\textbf{\sname achieves competitive in-domain performance and strong cross-domain generalization (synthetic $\rightarrow$ real).} We report accuracy and error metrics on the NYUv2 test set, comparing \sname{} against classical baselines and modern learning-based methods. Notably, \sname{} not only yields in-domain results comparable to DPT but also demonstrates competitive zero-shot generalization from synthetic to real data, matching the robustness of Depth Anything v2~\cite{yang2024depth2} and Marigold~\cite{ke2025marigold}.
}\label{tab:depth}
\vspace{-2pt}

  \centering\small
  \resizebox{\textwidth}{!}{%
  \begin{tabular}{@{}lllccccccccc@{}}
    \toprule
       \multirow{2}{*}{Method} &  \multirow{2}{*}{Pre-training}& \multirow{2}{*}{Training}
              & \multicolumn{3}{c}{Depth Accuracy} 
          & \multicolumn{3}{c}{Depth Error} \\
    \cmidrule(lr){4-6}\cmidrule(lr){7-9}
      & &
           & $\delta \!>\!1.25$ $\uparrow$ & $\delta \!>\!1.25^{2}$ $\uparrow$ & $\delta \!>\!1.25^{3}$  $\uparrow$
           & AbsRel $\downarrow$ & RMSE $\downarrow$  & log10 $\downarrow$ \\           
    \midrule
    \rowcolor{verylightgray}
   \multicolumn{9}{c}{\textit{\textbf{In-domain depth estimation (trained on NYUv2)}}} \\
     DPT & IN-1K & NYUv2 
                                 & 0.839 & 0.971 &  \textbf{0.992}
                                 &  0.132 & 0.457 & 0.055  \\
     \sname(ours)                & IN-1K & NYUv2 
                                  & \textbf{0.846} & \textbf{0.972} & \textbf{0.992} 
                                 & \textbf{0.130} & \textbf{0.451} & \textbf{0.054} \\
    \midrule
        \rowcolor{verylightgray}
        \multicolumn{9}{c}{\textit{\textbf{Joint segmentation--depth learning}}} \\
    Mousavian et al.~\cite{mousavian2016joint} & IN-1K & NYUv2 & 0.568 & 0.856 &  0.956 &  0.200 & 0.816 & 0.061  \\
    Simsar et al.~\cite{simsar2022object} & IN-1K & NYUv2  & 0.847 & 0.971 &  \textbf{0.993} &  \textbf{0.116} & 0.448 & -  \\
    \sname(ours)                & IN-1K & NYUv2 
                                  & \textbf{0.855} & \textbf{0.974} & \textbf{0.993} 
                                 & 0.123 & \textbf{0.433} & \textbf{0.052} \\
    \midrule
    \rowcolor{verylightgray}
\multicolumn{9}{c}{\textit{\textbf{Cross-domain zero-shot depth estimation (HyperSim $\rightarrow$ NYUv2)}}} \\
    Marigold~\cite{ke2025marigold}  & Laion-5b & HyperSim 
                                 &  0.375 & 0.659  &  0.833
                                 &  \textbf{0.542} & 1.243 &  0.171 \\
    Depth Anything v2~\cite{yang2024depth2} & LVM-142M & HyperSim 
                                 &  0.592 & \textbf{0.902} &  \textbf{0.960}
                                 &  0.749 & 0.808 &  0.110 \\
     \sname(ours)                & IN-1K & HyperSim 
                                  & \textbf{0.632} & 0.892 & \textbf{0.960} 
                                 & 0.583 & \textbf{0.740} & \textbf{0.102} \\
                                 
    \bottomrule
  \end{tabular}}
  \vspace{-5pt}
\end{table*}

\begin{table*}[t]
\caption{
\textbf{Ablation study of hierarchical encoding and decoding in \sname{}.}
We compare the full model with a variant that removes the reverse hierarchy in the decoder on the NYUv2 test set. The results highlight the role of coarse-to-fine unpooling in depth prediction.}\label{tab:ablation}
\vspace{-5pt}
  
  \centering\small
  \begin{tabular}{@{}lccccccccc@{}}
    \toprule
       \multirow{2}{*}{Method}
              & \multicolumn{3}{c}{Depth Accuracy} 
          & \multicolumn{3}{c}{Depth Error} \\
    \cmidrule(lr){2-4}\cmidrule(lr){5-7}
           & $\delta \!>\!1.25$ $\uparrow$ & $\delta \!>\!1.25^{2}$ $\uparrow$ & $\delta \!>\!1.25^{3}$  $\uparrow$
           & AbsRel $\downarrow$ & RMSE $\downarrow$  & log10 $\downarrow$ \\           
    \midrule
     DPT & 0.839 & 0.971 &  \textbf{0.992} &  0.132 & 0.457 & 0.055  \\
     \sname with forward hierarchy only & 0.755 & 0.951 &  0.989 &  0.163 & 0.552 & 0.069  \\
     \sname(ours) & \textbf{0.846} & \textbf{0.972} & \textbf{0.992} & \textbf{0.130} & \textbf{0.451} & \textbf{0.054} \\
                                 
    \bottomrule
  \end{tabular}
  \vspace{-5pt}
\end{table*}

\begin{figure*}[t]
  \centering\footnotesize
  \setlength{\tabcolsep}{0.5pt}   
  \renewcommand{\arraystretch}{1.0}
  \begin{tabular}{@{} ccc ccc ccc@{}}
    Image & Segment & Depth & 
    \multicolumn{3}{c}{\sname: 64, 32, 16 segments} &
    \multicolumn{3}{c}{CAST: 64, 32, 16 segments} \\[1pt]

    \includegraphics[width=0.108\textwidth]{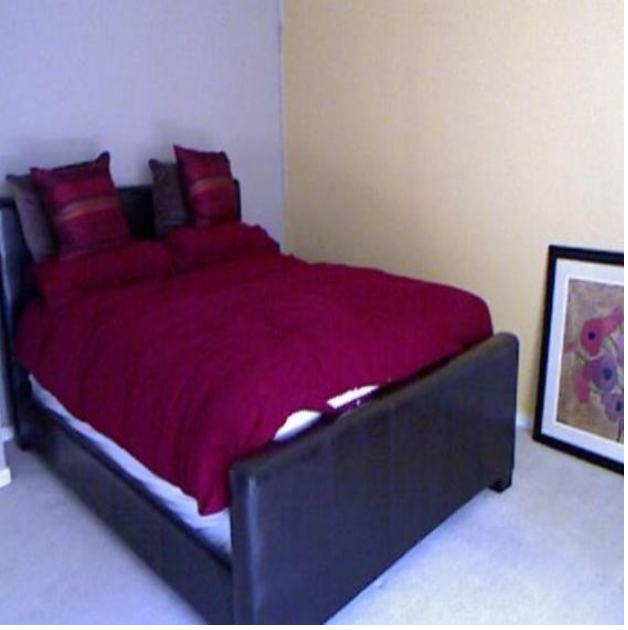} &
    \includegraphics[width=0.108\textwidth]{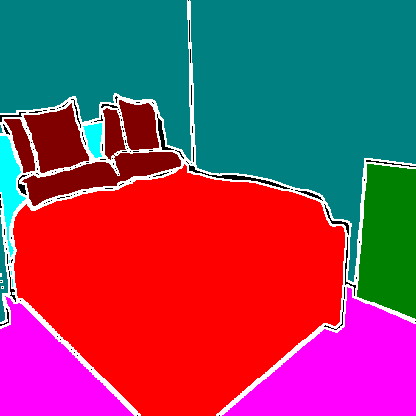} &
    \includegraphics[width=0.108\textwidth]{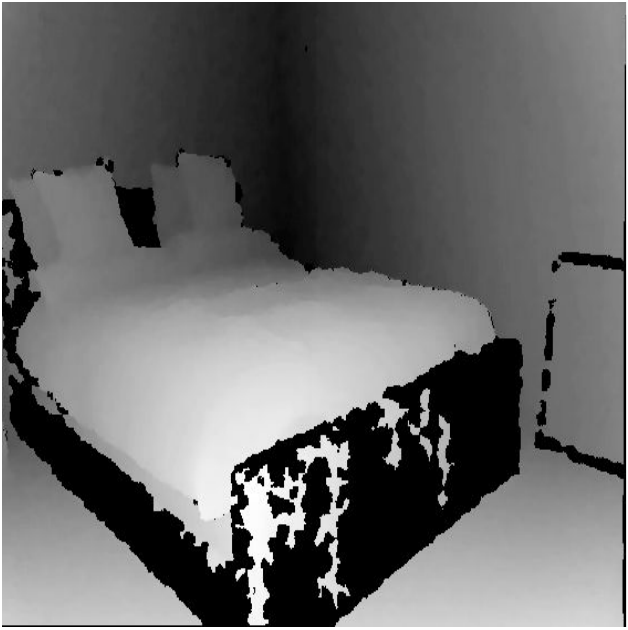} &
        
    \includegraphics[width=0.108\textwidth]{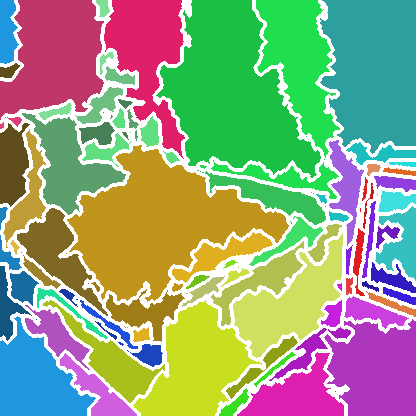} &
    \includegraphics[width=0.108\textwidth]{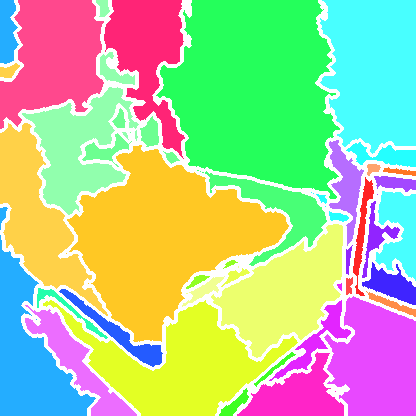} &
    \includegraphics[width=0.108\textwidth]{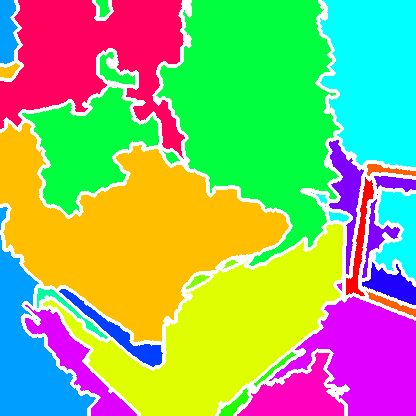} &

    \includegraphics[width=0.108\textwidth]{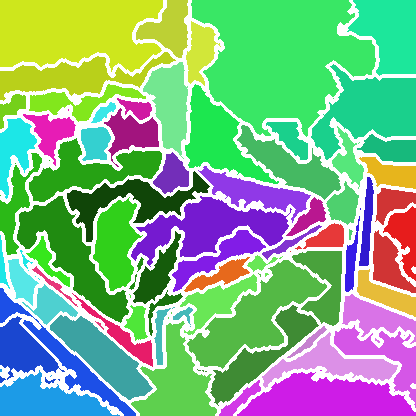} &
    \includegraphics[width=0.108\textwidth]{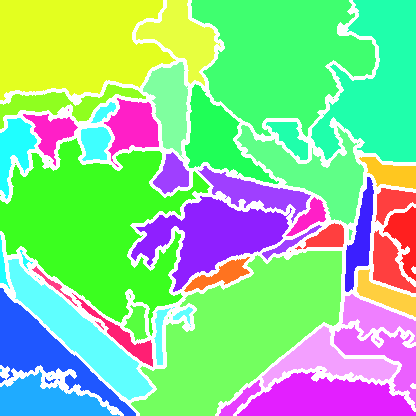} &
    \includegraphics[width=0.108\textwidth]{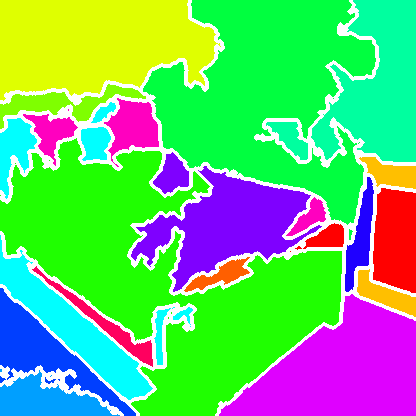} \\

    \includegraphics[width=0.108\textwidth]{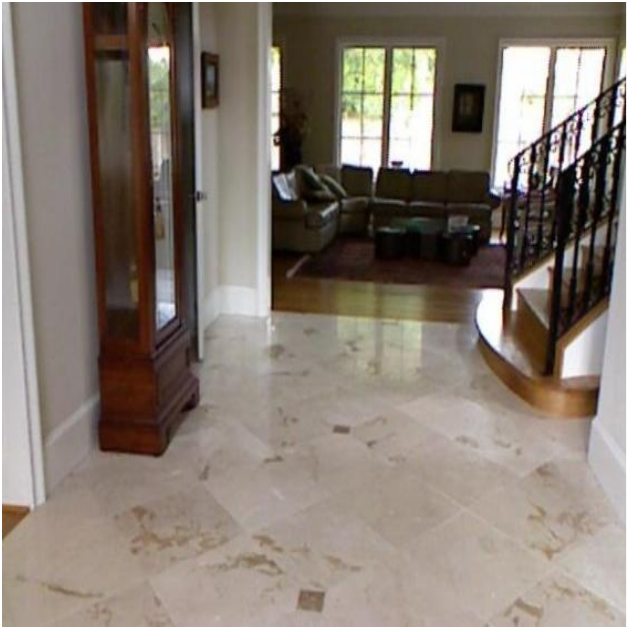} &
    \includegraphics[width=0.108\textwidth]{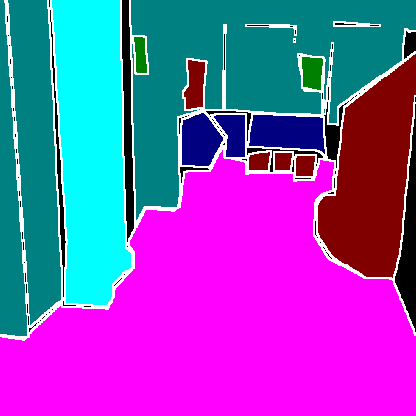} &
    \includegraphics[width=0.108\textwidth]{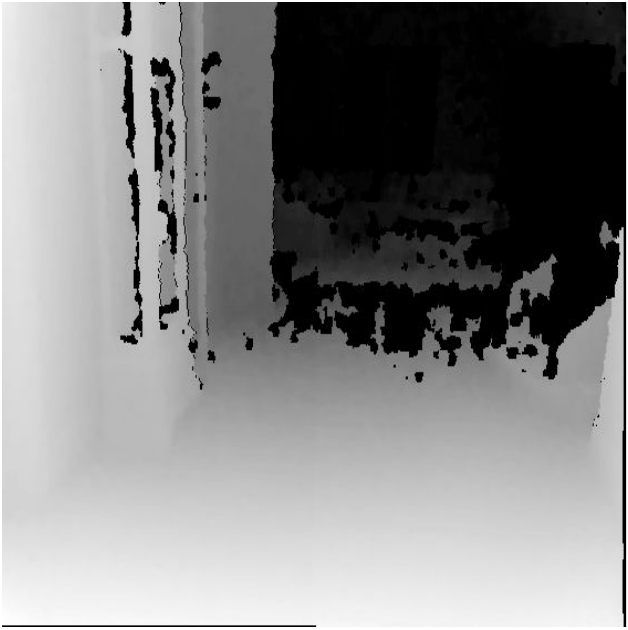} &

    \includegraphics[width=0.108\textwidth]{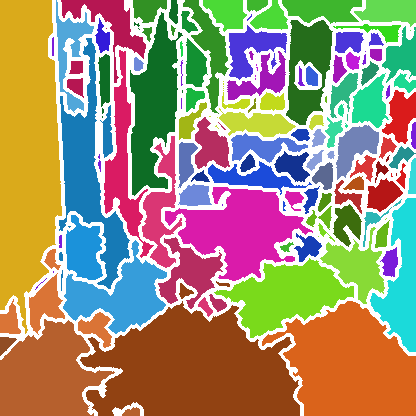} & 
    \includegraphics[width=0.108\textwidth]{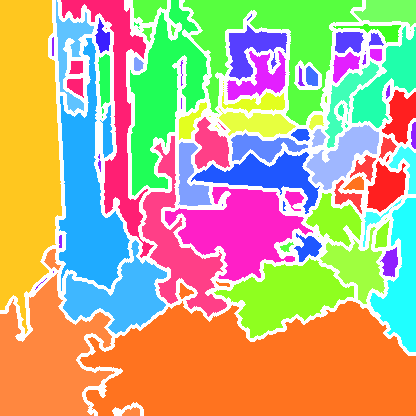} &
     \includegraphics[width=0.108\textwidth]{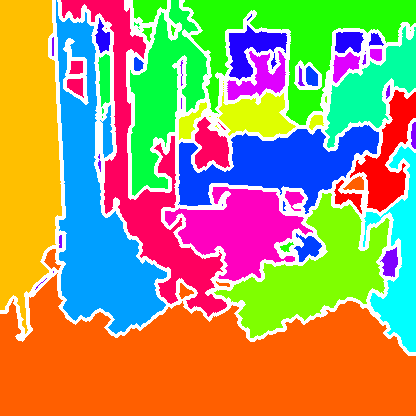} &

    \includegraphics[width=0.108\textwidth]{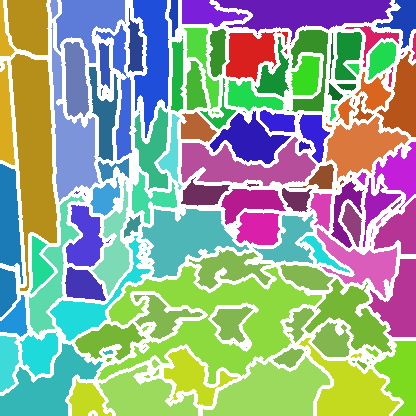} &
    \includegraphics[width=0.108\textwidth]{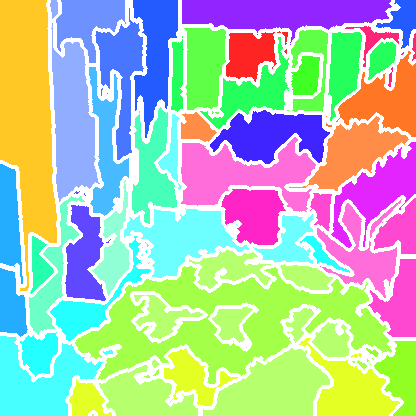} &
    \includegraphics[width=0.108\textwidth]{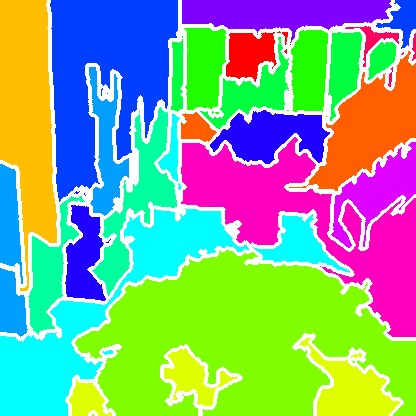} \\
      
    \end{tabular} 
  \vspace{-4pt}
  \caption{
\textbf{\sname learns depth-aware segment hierarchies, while CAST relies on visual cues.}
We compare segmentations from \sname and CAST~\citep{ke2024learning} at the same hierarchy levels: 64, 32, and 16 segments. \sname captures meaningful part structures, such as separating the blanket and pillow from the bed (row 1). It also decomposes large structures like the floor based on depth, grouping nearby regions into a single large segment while dividing distant areas into smaller ones (row 2). In contrast, CAST relies on appearance cues and fails to capture geometric structure. For instance, it groups white floor regions by color but divides them arbitrarily, ignoring depth. These results highlight the value of depth supervision in learning 3D-aware segmentations.
  }\label{fig:cast}
    \vspace{-5pt}

\end{figure*}

\textbf{Architecture.}
With graph pooling and unpooling layers, the encoder consists of three stages, each with two ViT blocks followed by graph pooling, progressively reducing the number of segment tokens to 256, 128, and 64. The decoder mirrors this with unpooling and receives skip connections from the corresponding encoder stages.

We train \sname and DPT on NYUv2 using a batch size of 16 for 50 epochs with the Adam optimizer~\citep{kingma2014adam} and a learning rate of 5e-5. With pretrained ResNet and ViT backbones, we follow DPT’s default training recipe, including the scale-invariant logarithmic loss computed against ground-truth depth. At inference time, predicted depth maps at 384$\times$384 resolution are bilinearly upsampled to 640$\times$480 to match the ground-truth size.

\vspace{-5pt}

\subsection{Segment-consistent depth estimation}
\vspace{-5pt}

\sname generates structured depth maps using a learned segment hierarchy. We first visualize this structural alignment, then evaluate boundary quality and coherence, and finally demonstrate that this hierarchical approach improves per-pixel metrics and efficiency. \cref{fig-main} shows how the segment hierarchy in \sname benefits geometry. The learned segments capture contours of objects, such as desks in a classroom, allowing the depth to clearly separate them from the floor. They also decompose larger structures, like tables, into parts, enabling smooth depth transitions from front to back.
This suggests that structure guides depth prediction toward more accurate and interpretable results.

\textbf{Boundary accuracy.}
We assess the structural quality of \sname by comparing occlusion boundaries against DPT on the NYUv2-OC++ dataset~\citep{ramamonjisoa2020predicting}. \cref{fig:boundary} shows predicted depth maps and their occlusion boundaries, extracted using a Canny edge detector~\citep{canny1986computational}. For quantitative evaluation, we follow the standard protocol~\citep{koch2018evaluation} and compute the average Chamfer distance~\citep{fan2017point} in two directions: from prediction to ground truth, and vice versa.
\sname produces sharper contours and outperforms DPT on both metrics with its fine-grained segmentation.

\textbf{Intra-segment coherence.}
Beyond boundary, we evaluate how coherently depth values vary within each segment. We employ object-wise depth accuracy and error, treating ground-truth semantic segmentation masks as structural references on NYUv2~\citep{Silberman:ECCV12}. As shown in \cref{fig:boundary}, \sname produces smoother depth variations within segments. This is reflected quantitatively in \cref{tab:boundary}. \sname achieves lower error and higher accuracy within segments, confirming that our hierarchical pooling promotes smoother and more coherent depth predictions.

\textbf{Per-pixel metrics \& generalization.} 
\cref{tab:depth} reports standard per-pixel depth metrics under both in-domain and cross-domain settings. 
nder in-domain evaluation on NYUv2, \sname{} outperforms the DPT baseline and prior approaches that incorporate segmentation cues for depth estimation.
For example, methods such as~\cite{simsar2022object} enforce object-level consistency via over-segmentation, while multi-task frameworks like~\cite{mousavian2016joint} jointly supervise semantics and depth. The benefits of structural modeling become pronounced under cross-domain evaluation.
Despite relying only on ImageNet pre-training, \sname{} exhibits strong zero-shot generalization from HyperSim to NYUv2, outperforming large-scale foundation models on most metrics.
In particular, it surpasses Depth Anything v2~\citep{yang2024depth2}, which leverages a DINOv2~\citep{oquab2023dinov2} encoder, and Marigold~\citep{ke2025marigold}, pre-trained on billions of images.
These results suggest that incorporating explicit structural constraints offers a data-efficient path to robust depth estimation across domains. We report results on a diverse set of benchmarks in \cref{appx:diverse}.

\begin{figure*}[t]
  \centering\footnotesize
  \setlength{\tabcolsep}{10pt}   
  \renewcommand{\arraystretch}{1.0}

    \newcommand{\overlayimage}[3]{%
    \tikz[baseline]{
      \node[inner sep=0pt] (img) {\includegraphics[trim=5 12.4 5 12.4, clip, width=#1]{#2}};
      \node[anchor=south east, font=\footnotesize, fill=white, text opacity=0.8, draw=black, rounded corners=1pt, xshift=-2pt, yshift=2pt] 
        at (img.south east) {#3};
    }}

  \newcommand{\hspacesmall}{\hspace{1.5pt}}
  \newcommand{\hspacelarge}{\hspace{2.5pt}}
  \newcommand{\hspacelarger}{\hspace{5pt}}
 
  \begin{tabular}{
  @{}
  c@{\hspacelarge}
  c@{\hspacesmall}c@{\hspacesmall}c@{\hspacelarger}
  c@{\hspacelarger}c@{\hspacelarger}c@{}
  }
    Image &
    DPT & \sname (ours) & GT &
    DPT & \sname (ours) & GT \\[1pt]

    \includegraphics[width=0.13\textwidth]{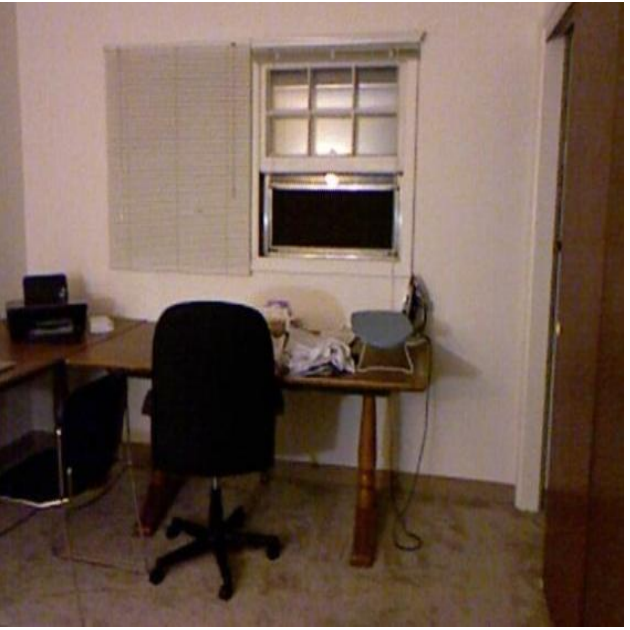} &
    \includegraphics[trim=5 12.4 5 12.4, clip, width=0.135\textwidth]{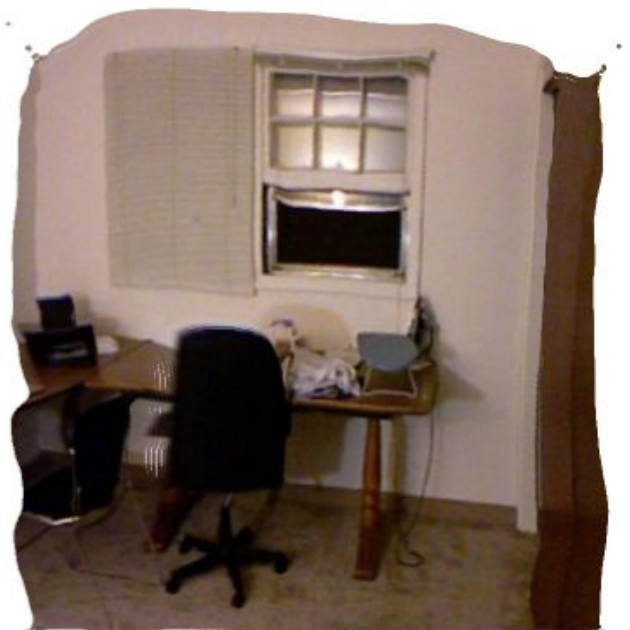} &
    \includegraphics[trim=5 12.4 5 12.4, clip, width=0.135\textwidth]{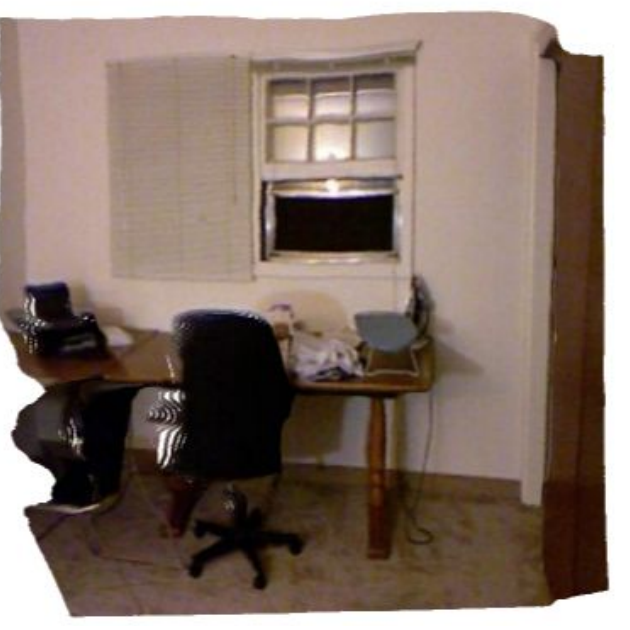} &
    \includegraphics[trim=5 12.4 5 12.4, clip, width=0.135\textwidth]{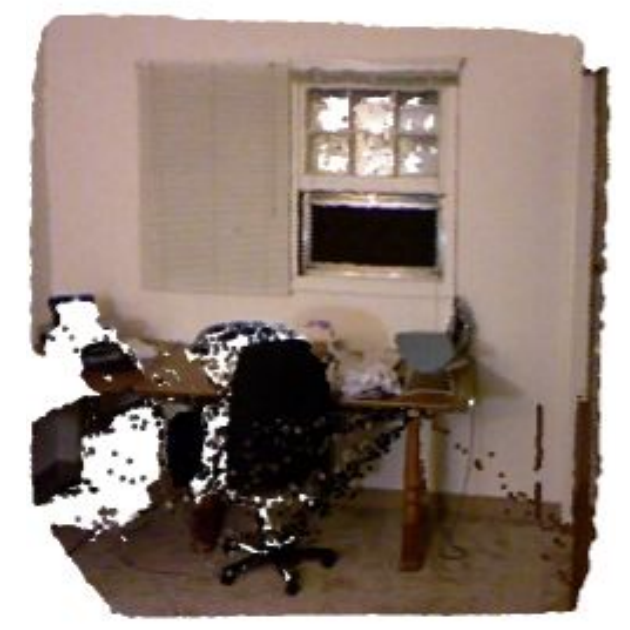} & 
    \includegraphics[width=0.135\textwidth]{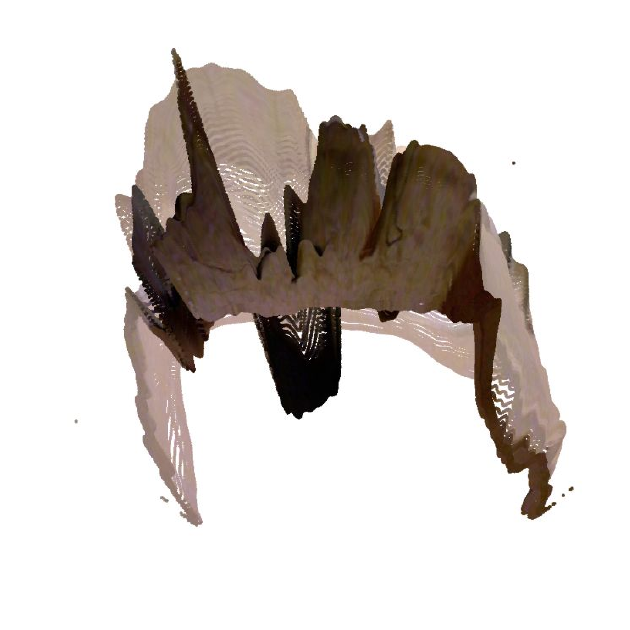} &
    \includegraphics[trim=5 12.4 5 12.4, clip, width=0.135\textwidth]{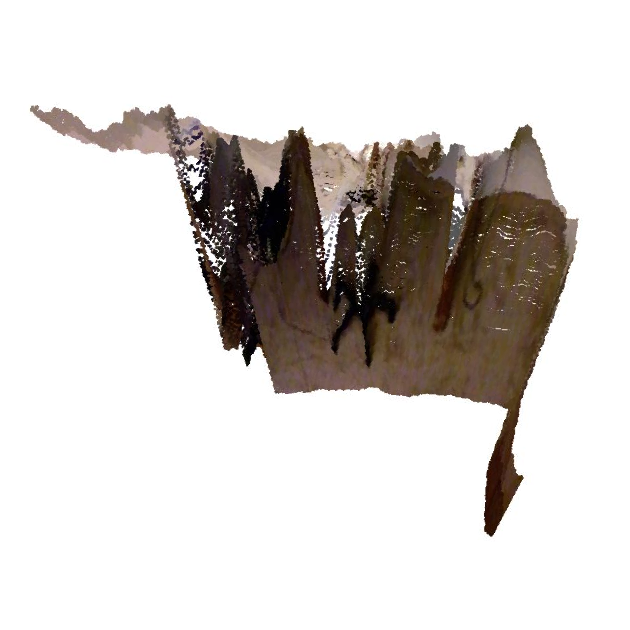} &
    \includegraphics[width=0.135\textwidth]{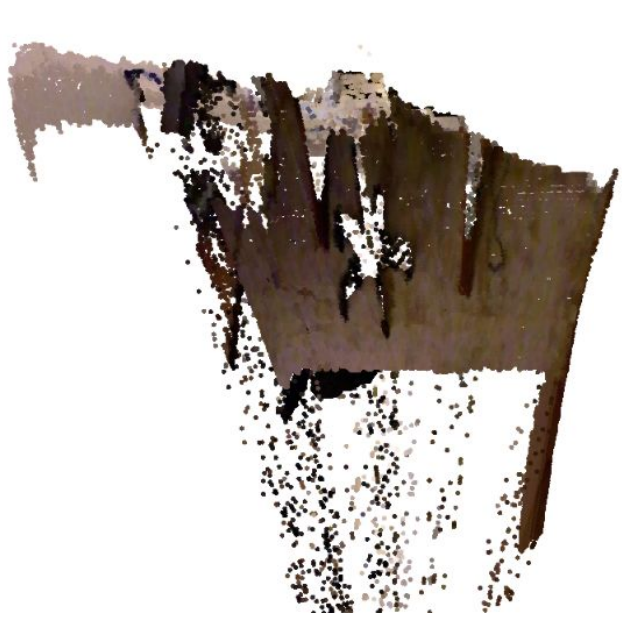}
     \\

  \end{tabular}
    \caption{%
\textbf{\sname produces structured 3D reconstructions.}  
We visualize 3D point clouds reconstructed from single-view depth maps, following the semantic scene completion protocol~\citep{song2017semantic}, using predictions from DPT, \sname, and the ground truth on NYUv2 examples.  
Frontal views (cols 2-4) show that DPT fails to preserve planar structures, producing curved wall boundaries, whereas \sname more accurately recovers straight lines.  
This difference is more apparent in the bird’s-eye views (cols 5-7): DPT yields warped surfaces, while \sname produces flatter layouts.
}
    \label{fig:point}

\vspace{-12pt}

\end{figure*}
\begin{figure*}[t]
\begin{minipage}[t]{0.4\textwidth}
\vspace{-41pt}
\centering\scriptsize
\captionof{table}{%
\textbf{3D alignment induced by structured depth estimation.}
We compute the Chamfer distance between point clouds from the predicted and ground-truth depths. \sname achieves lower errors than DPT, indicating improved geometric alignment.
}\label{tab:3d}
\resizebox{0.8\textwidth}{!}{\begin{tabular}{@{}lcc@{}}
      \toprule
      Method & Precision / Recall $\downarrow$ \\
      \midrule
      DPT & 0.171 / 0.251 \\
      \sname (ours) & \textbf{0.158} / \textbf{0.244} \\
      \bottomrule
\end{tabular}}
\end{minipage}%
\hspace{.018\textwidth}
\begin{minipage}[t]{0.59\textwidth}
  \centering\footnotesize
  \setlength{\tabcolsep}{1pt}   
  \renewcommand{\arraystretch}{1.0}
  \begin{tabular}{@{}ccc@{}}
    Image &
    Segment &
    3D Parts \\[1pt]
    \includegraphics[width=0.33\linewidth]{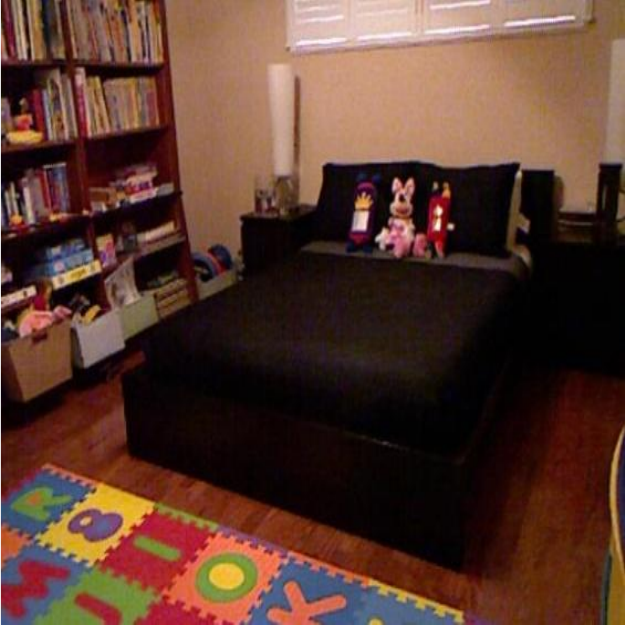} &
    \includegraphics[width=0.33\linewidth]{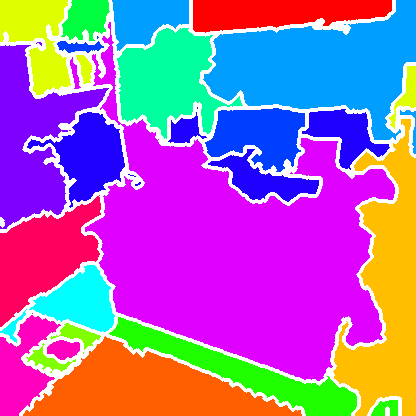} &
    \includegraphics[width=0.33\linewidth]{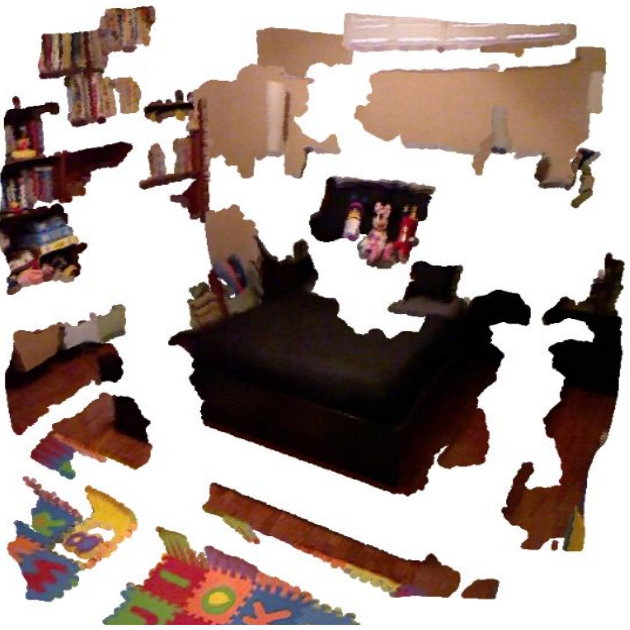}
  \end{tabular}
  \caption{
\textbf{\sname discovers 3D part structures.}
Concurrent segmentation and depth estimation enable part-level decomposition of the 3D point clouds.
  }
  \label{fig:3d}
  \end{minipage}
  \vspace{-5pt}
\end{figure*}

\textbf{Computational efficiency.} For a fair and hardware-independent comparison of structural complexity, we report both the number of parameters and the computational cost in FLOPs. DPT-Hybrid contains 41.88 million parameters and requires 135.0 GFLOPs. In contrast, SHED uses 56.58 million parameters but reduces the computation to 103.2 GFLOPs, which is approximately a \textbf{24\% decrease in FLOPs}. This substantial reduction demonstrates that SHED is structurally more efficient and achieves lower theoretical latency despite having a slightly larger parameter count.

\textbf{Ablation.} We conducted an ablation study to isolate the contributions of the hierarchical clustering in the encoder and the hierarchy reversing in the decoder. To verify the necessity of our proposed decoder, we experimented with a variant of SHED by removing the progressive unpooling process in the decoder. Table~\ref{tab:ablation} shows removing the reverse hierarchy leads to a sharp performance degradation, falling significantly behind the DPT baseline. This demonstrates that the coarse-to-fine unpooling mechanism in the decoder is essential to recover fine-grained spatial details from the grouped representations.

\vspace{-5pt}

\subsection{Structure-aware representation learning}

\begin{table}[t]
  \centering
  \small
  \caption{\textbf{Semantic segmentation comparison results on ADE20K.} Segment hierarchies learned by \sname{} improve alignment with semantic regions and boundaries on ADE20K.}
  \label{tab:cast}
  \vspace{-4pt}
  \begin{tabular}{@{}lcc@{}}
    \toprule
    Method & mIoU & boundary F-score \\
    \midrule
    CAST & 43.1 & 36.5 \\
    SHED (ours) & \textbf{44.5} & \textbf{37.7} \\
    \bottomrule
  \end{tabular}
  \vspace{-15pt}
\end{table}

\textbf{Depth-aware image segmentation.} We analyze the segment hierarchy learned by \sname by comparing it to CAST, an encoder trained for image recognition using segment-based representations. Specifically, we use CAST, trained on ImageNet~\citep{deng2009imagenet} with the MoCo-v3 objective~\citep{chen2021empirical}, a self-supervised learning by instance discrimination~\citep{wu2018unsupervised} that clusters visually similar images. To ensure a fair comparison, we adapt the graph pooling layers of \sname to generate the same number of segments (64, 32, and 16) from 196 superpixels at a 224$\times$224 resolution. 

\cref{fig:cast} shows that \sname learns hierarchical structures that align with scene geometry. It separates objects like blankets and decomposes large structures such as floors into segments that reflect their spatial extent. In contrast, CAST groups regions based on appearance. For example, it clusters white floor areas by color but fails to account for geometric cues. We attribute this difference to the training objective. CAST learns segments through image-level recognition, while \sname is guided by dense prediction.

\textbf{Semantic segmentation.} We further evaluate whether the learned segment hierarchies align with semantic structures on the ADE20K dataset~\citep{zhou2017scene}.
Following the CAST evaluation protocol, we report region-level mean IoU and boundary F-score between the predicted segments and ground-truth annotations.
As shown in Table~\ref{tab:cast}, \sname{} consistently outperforms CAST on both metrics. Notably, this improvement cannot be attributed solely to encoder pre-training.
While both models leverage segment hierarchy, \sname{} incorporates a hierarchical decoding pipeline that progressively reverses the segment hierarchy, enabling fine-grained spatial details to be recovered from grouped representations.
This coarse-to-fine decoding process facilitates more accurate alignment with semantic boundaries, leading to improved segmentation quality without requiring explicit semantic supervision.


\vspace{-5pt}
\subsection{3D scene reconstruction with part structures}

We demonstrate SHED’s capability for 3D scene understanding. While plausible pixel values may suffice for 2D depth estimation, accurate and structured depth is particularly critical when projected into 3D space. Accordingly, \sname enables high-quality 3D reconstruction and supports unsupervised 3D part discovery through segmentation. To evaluate the structural quality of depth maps, we project them into 3D point clouds on the NYUv2 dataset~\citep{Silberman:ECCV12}, following the semantic scene completion protocol~\citep{song2017semantic} and using NYUv2 camera intrinsics. For interpretability, all depth values are scaled by 1/1000.
\cref{fig:point} shows that \sname produces cleaner reconstructions with sharper boundaries and flatter surfaces that better align with ground truth geometry, whereas DPT yields curvier, less faithful shapes.
We quantify reconstruction performance with the Chamfer distance~\citep{fan2017point} in both directions. \cref{tab:3d} shows that \sname consistently achieves lower distances than DPT, confirming its advantage in structured 3D prediction. By jointly predicting segmentation and depth, \sname lifts 2D parts into 3D space, enabling part-level decomposition of scenes. 

\cref{fig:3d} shows an example from NYUv2, where segments corresponding to objects form coherent 3D structures in point clouds. This demonstrates SHED’s potential for unsupervised 3D part reasoning, a key capability for interactive and dynamic scene understanding~\citep{mo2019partnet}.


\section{Conclusion}
We shed light on the role of segmentation in depth estimation. \sname learns a segment hierarchy in the encoder and reverses it in the decoder to predict dense maps. This results in depth maps with segment-consistent structure, structure-aware representations, and coherent 3D scenes with interpretable parts. Our principle of unifying reconstruction and reorganization offers a new direction for 3D vision and robotics, particularly for tasks that require fine-grained interaction with physical components. Additional results and limitations are discussed in \cref{appx:limitation}.

\textbf{Impact statement.} This research was conducted responsibly based on the principles outlined in the ICML Code of Ethics. This technology can enhance the 3D environmental perception of autonomous driving systems, thereby improving road safety, and can help robots interact more safely and efficiently with their surroundings. Before deploying this model in real-world scenarios, it must undergo rigorous and thorough validation for robustness and safety across a wide range of conditions.



\bibliography{example_paper}
\bibliographystyle{icml2026}

\newpage
\appendix
\onecolumn


\section{Implementation Details}
\label{appx:detail}

\subsection{Training details}

We train our model on the NYUv2 dataset~\citep{Silberman:ECCV12} using the official training split. Each RGB image is first cropped to remove invalid boundaries (coordinates: 43, 45, 608, 472), then resized to $384 \times 384$ resolution. The corresponding depth maps undergo the same spatial preprocessing and are normalized by dividing raw depth values by 1000.

For data augmentation, we apply horizontal flipping with a probability of 0.5, gamma correction with $\gamma \in [0.9, 1.1]$, brightness scaling using a random factor from $[0.75, 1.25]$, and per-channel color jittering with multiplicative factors in $[0.9, 1.1]$. After augmentation, random spatial crops of size $384 \times 384$ are applied to both images and depth maps.

For tokenization, we generate superpixels using OpenCV’s SEEDS~\citep{van2012seeds} algorithm. Each image is segmented into 676 superpixels using a single-level hierarchy (\texttt{num\_levels=1}) and a histogram bin size of 5. The algorithm is run for 50 iterations to refine superpixel boundaries.









\subsection{Evaluation details}

We evaluate on the official NYUv2~\citep{Silberman:ECCV12} test split, which contains 654 images. All evaluations use an input resolution of 384$\times$384 pixels, with depth values clamped to the range $[10^{-3}, 10.0]$.

\noindent\textbf{Per-pixel depth metrics.}
We compute standard depth estimation metrics over valid pixels where ground truth depth is available. Depth error metrics include AbsRel (mean absolute relative error), RMSE (root mean squared error), and Log10 (mean absolute logarithmic error). Accuracy is measured using threshold metrics $\delta^1$, $\delta^2$, and $\delta^3$, which denote the percentage of pixels where the predicted-to-ground-truth depth ratio is below $1.25$, $1.25^2$, and $1.25^3$, respectively. All metrics are computed with numerical safeguards, including epsilon clamping at 1e-6 to prevent division by zero and log-domain errors.

\noindent\textbf{Occlusion boundary.} 
We follow
the evaluation protocol of the NYUv2-OC++ dataset~\citep{ramamonjisoa2020predicting} to assess occlusion boundary accuracy. Each predicted depth map is first min-max normalized, followed by the application of the OpenCV Canny edge detector~\citep{canny1986computational} with low and high thresholds of 100 and 200 to produce a binary mask of predicted boundary pixels. Using the ground-truth boundary labels from NYUv2-OC++, we compute two metrics: $\varepsilon_a$ (accuracy), the average distance from each predicted edge pixel to the nearest ground-truth edge; and $\varepsilon_c$ (consistency), the average distance from each ground-truth edge pixel to the nearest predicted edge. Both are reported in squared pixels, where lower values indicate better alignment, and 0 denotes perfect correspondence.

\noindent\textbf{3D scene reconstruction.}
We reconstruct 3D point clouds by back-projecting each pixel of the predicted depth maps into 3D space using the known camera intrinsics from the NYUv2 dataset, following the standard protocol for semantic scene completion~\citep{song2017semantic}. To evaluate reconstruction quality, we compute the Chamfer distance between the predicted and ground truth point clouds. Recall quantifies how well the predicted points cover the ground truth surface, while precision measures how accurately the predicted points align with the true geometry.

Detailed evaluation protocol for  \textbf{layout-aware retrieval} is provided in \cref{appx:retrieval}, respectively.
\section{Layout-aware Retrieval}
\label{appx:retrieval}

Our architecture not only improves depth prediction but also facilitates structure-aware representation learning.
First, \sname learns features that reflect scene layout, enabling more accurate layout-aware image retrieval than DPT~\citep{ranftl2021vision}.
Second, its segment hierarchy captures geometric cues informed by depth supervision, whereas CAST~\citep{ke2024learning} relies on visual cues.

\begin{figure*}[h]
  \centering\footnotesize
  \setlength{\tabcolsep}{0.5pt}
  \renewcommand{\arraystretch}{1.0}

\renewcommand{\thesubfigure}{\alph{subfigure})}
\captionsetup[sub]{labelformat=simple, labelsep=space}

\begin{minipage}{0.66\textwidth}
    \subcaption{Top-5 retrieval examples}
    \label{fig:retrival-a}
    \centering\small

    \newcommand{\retrievalheight}{40pt}
    \newcommand{\retrievalimg}[1]{\includegraphics[height=\retrievalheight]{fig/retrieval/#1.jpg}}
\newcommand{\retrievallabel}[1]{%
  \hspace{5pt}%
  \makebox[0pt][r]{%
    \vbox to \retrievalheight{%
      \vfil
      \hbox{\rotatebox[origin=c]{270}{\raisebox{0pt}[1ex][1ex]{\scriptsize #1}}}
      \vfil
    }%
  }
}

    \scriptsize
    \begin{tabular}{@{}cccccccc@{}}    
        \retrievalimg{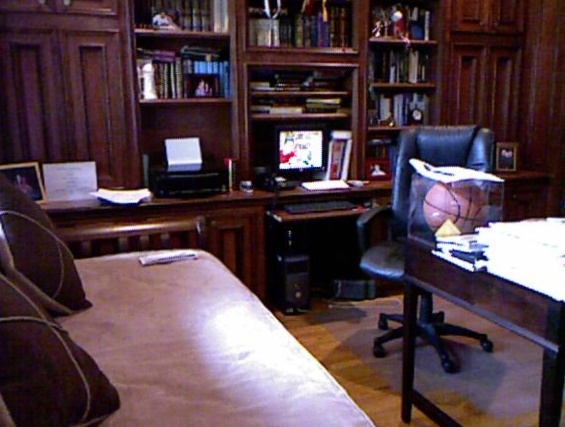} & \retrievalimg{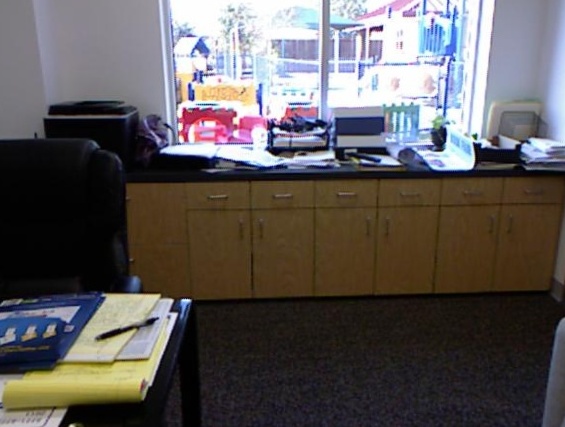} & \retrievalimg{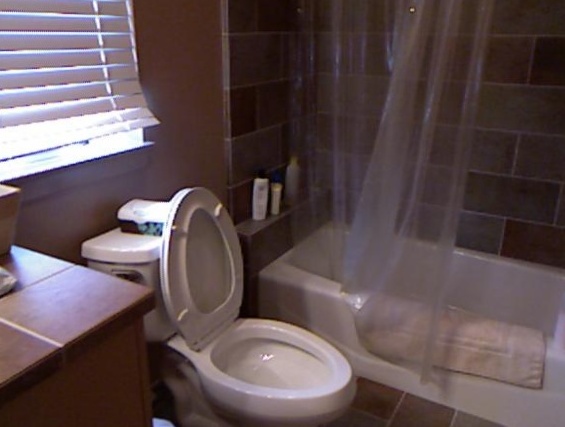} & \retrievalimg{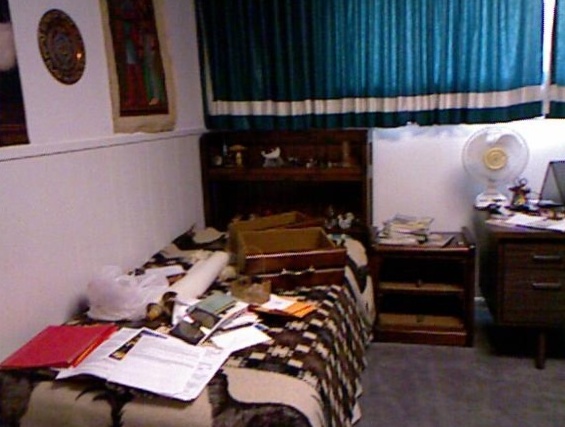} & \retrievalimg{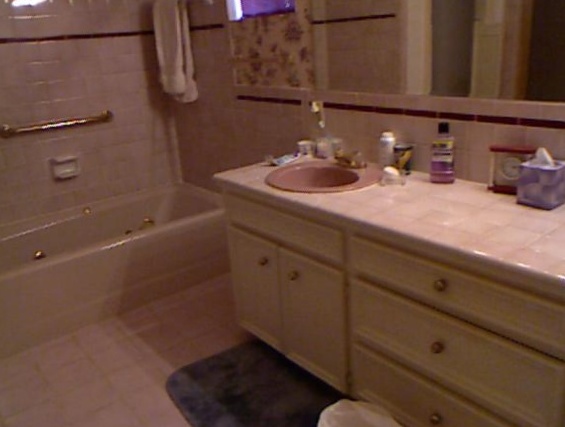} & \retrievalimg{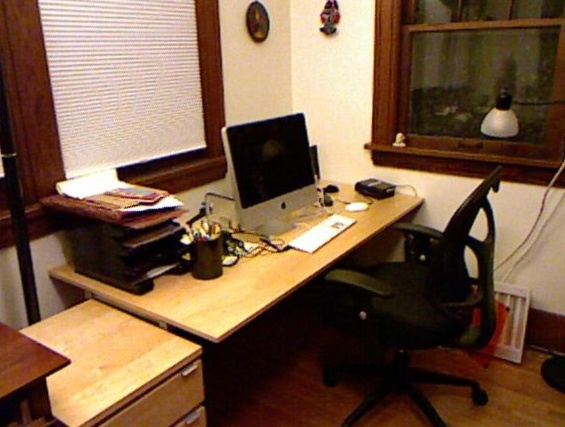}
        & \retrievallabel{DPT} \\
        \noalign{\vskip -2.0pt}
         Query & 0.713 & 0.638 & 0.624 & 0.621 & 0.617 \\    
        \noalign{\vskip -1pt}
        \retrievalimg{474} &
        \retrievalimg{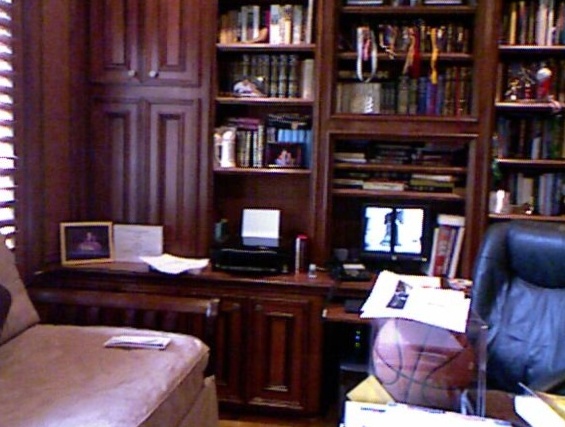} & \retrievalimg{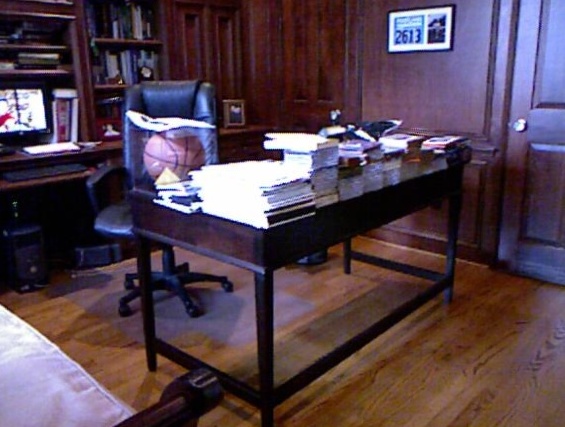} & \retrievalimg{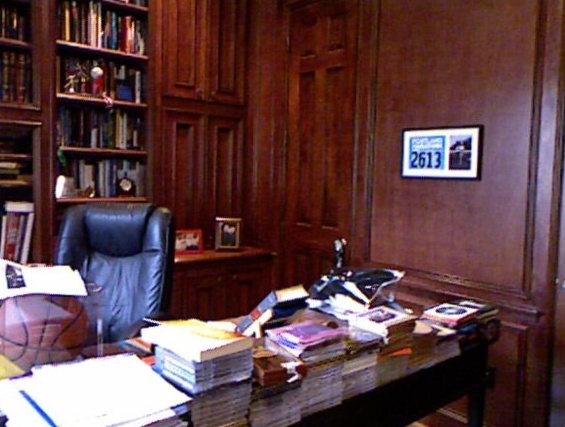} & \retrievalimg{357} & \retrievalimg{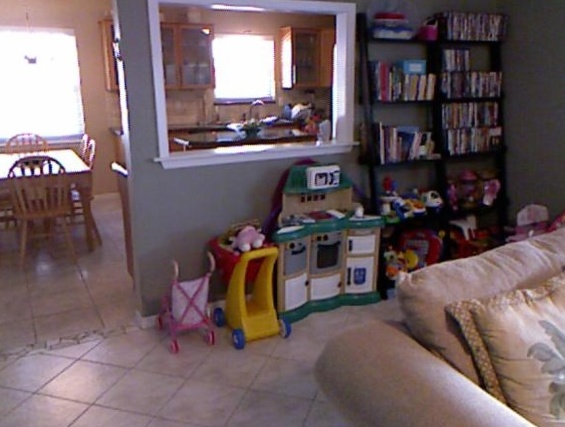}
        & \retrievallabel{\sname (ours)} \\
        \noalign{\vskip -2.0pt}
         Query & 0.885 & 0.844 & 0.812 & 0.792 & 0.787 \\   
    \end{tabular}
    
\end{minipage}
~~~~~
  \begin{minipage}{0.30\textwidth}
   \centering\scriptsize
   \subcaption{Top-$K$ retrieval accuracy (\%)\label{fig:retrieval_combined:a}}
   \label{tab:retrival-b}

    \setlength{\tabcolsep}{4pt}

  \resizebox{\textwidth}{!}{\begin{tabular}{lccc}
    \toprule
    Method & Top-1 & Top-3 & Top-5 \\
    \midrule
    \addlinespace[2pt]
    \multicolumn{4}{c}{\cellcolor{gray!20} \textit{Scene retrieval}} \\
    \addlinespace[2pt]
    DPT & 45.2 & 69.7 & 77.2 \\
    \sname (ours) & \textbf{60.5} & \textbf{78.7} & \textbf{87.0} \\
    \midrule
    \addlinespace[2pt]
    \multicolumn{4}{c}{\cellcolor{gray!20} \textit{Frame retrieval ($k$=5)}} \\
    \addlinespace[2pt]
    DPT & 18.5 & 31.0 & 38.3 \\
    \sname (ours) & \textbf{30.5} & \textbf{42.3} & \textbf{48.3} \\
    
    \bottomrule
  \end{tabular}}
\end{minipage}
  \caption{
\textbf{SHED learns layout-aware representations through depth supervision.}
We evaluate image retrieval on NYUv2 based on cosine similarity between class tokens from the final ViT block.  
\textbf{a)} Top-5 results (ranked left to right), with similarity scores shown below. \sname retrieves images with similar layouts, such as a central desk and a rear bookshelf, while DPT retrieves unrelated scenes.  
\textbf{b)} Top-$K$ accuracy at the scene and frame level ($k=5$), where the targets are different views from the same scene or nearby frames.  
\sname significantly outperforms DPT in all settings, indicating that our depth-guided segmentation effectively encodes spatial layout.}
  \label{fig:retrieval}
\end{figure*}


We provide a detailed explanation of our proposed metric, \textit{layout-aware retrieval}, which evaluates the structural quality of learned representations by measuring how well the model retrieves frames from the same 3D scene in a video, either across the full sequence or within nearby frames.

To compute this metric, we extract the \texttt{[CLS]} token from the final layer of the vision transformer in DPT and \sname for each image, denoted as $\mathbf{z}_{\text{cls}} \in \mathbb{R}^{D}$, and apply $l_2$ normalization. We then construct a full pairwise similarity matrix, where each entry is computed as the cosine similarity between $l_2$-normalized image embeddings: $S_{ij} = \mathbf{z}_{\text{cls}}^{(i)} \cdot \mathbf{z}_{\text{cls}}^{(j)}$.

Retrieval performance is evaluated in two settings: \textit{scene-level} and \textit{frame-level}. In the scene-level setting, the goal is to retrieve other frames from the same annotated scene, testing the model’s ability to maintain structural consistency under varying viewpoints. In the frame-level setting, each frame is treated as an independent query, focusing on retrieving visually similar frames regardless of scene membership. We additionally define a frame-$k$ variant, where retrieval is restricted to the $k$ temporally adjacent frames, allowing us to assess the model’s sensitivity to local layout changes. We report top-$K$ nearest neighbor retrieval accuracy on the NYUv2~\citep{Silberman:ECCV12} test set.

\begin{figure*}[ht!]
  \centering
  \footnotesize
  \setlength{\tabcolsep}{1pt}

  \begin{subfigure}[b]{0.33\linewidth}
    \subcaption{Top-1}
    \includegraphics[width=\linewidth]{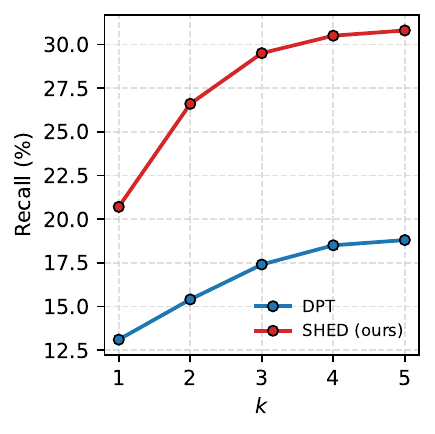}
    \label{fig:top1}
  \end{subfigure}\hfill
  \begin{subfigure}[b]{0.33\linewidth}
  \subcaption{Top-3}
    \includegraphics[width=\linewidth]{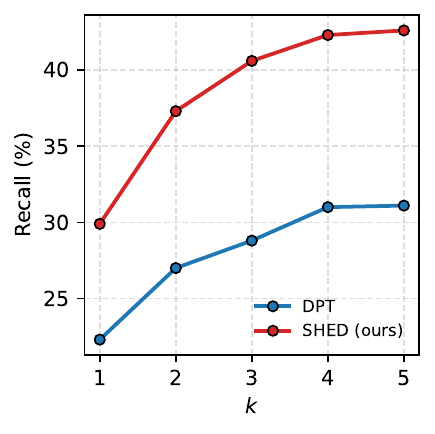}
    \label{fig:top3}
  \end{subfigure}\hfill
  \begin{subfigure}[b]{0.33\linewidth}
  \subcaption{Top-5}
    \includegraphics[width=\linewidth]{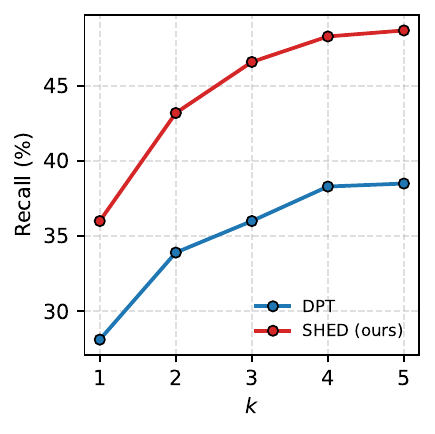}
    \label{fig:top5}
  \end{subfigure}

    \vspace{-11pt}
  \caption{\textbf{Frame-$k$ recall with varying temporal range $k$.}
           We report Top-$K$ retrieval recall (\(K \in \{1,3,5\}\)) for
           DPT and \sname{} across temporal ranges \(k \in [1,5]\).
           The consistent gains highlight the robustness of our
           depth-supervised representation to spatial and viewpoint
           changes.}
  \label{fig:layout}
\end{figure*}

To further evaluate the robustness of layout-aware retrieval, we vary both the candidate set size ($k$) and the top-$K$ threshold. A smaller $k$ imposes a stricter constraint, requiring the model to identify the most similar frame from a limited pool. As shown in \cref{fig:layout}, \sname consistently outperforms DPT across all settings. Performance improves for both methods as $k$ increases, with the largest gap observed at low top-$K$ values. These results suggest that \sname captures layout similarity more precisely and excels at retrieving the most relevant match.




\section{Evaluation on Diverse Benchmarks}
\label{appx:diverse}

we extend our experiments to a mixed-domain training setup using Virtual KITTI~\cite{cabon2020virtual}, HyperSim~\cite{roberts2021hypersim}, and MegaDepth~\cite{li2018megadepth}. We train the SHED using a uniform sampling strategy across all datasets for 20 epochs with batch size of 64. We then evaluate this model in a fully zero-shot manner on three unseen datasets, including KITTI (outdoor)~\cite{geiger2013vision}, NYUv2 (indoor)~\cite{Silberman:ECCV12}, SUN-RGBD (indoor)~\cite{song2015sun}.

The results below demonstrate that SHED consistently outperforms the DPT baseline across domains, including both indoor and outdoor scenes. The segment-hierarchy inductive bias of SHED is not specific to the NYUv2 domain but transfers effectively to outdoor and mixed-domain settings, consistently outperforming the baseline.

\begin{table*}[t]
\centering\small
\caption{
\textbf{\sname improves mixed data setting.} We evaluate standard depth accuracy and error metrics on more diverse dataset. \sname delivers competitive per-pixel depth estimation performance comparable to DPT when trained mixed training setting in cross-domain zero-shot evaluation.
}\label{tab:diverse}
\vspace{-5pt}
  \centering\small
  \begin{tabular}{@{}lccccccccc@{}}
    \toprule
       \multirow{2}{*}{Method}   & \multicolumn{2}{c}{KITTI}   & \multicolumn{2}{c}{NYUv2} & \multicolumn{2}{c}{SUN-RGBD}  \\
    \cmidrule(lr){2-3}\cmidrule(lr){4-5}\cmidrule{6-7} & AbsRel $\downarrow$ & $\delta \!>\!1.25$ $\uparrow$ &   AbsRel $\downarrow$  & $\delta \!>\!1.25$ $\uparrow$ &  AbsRel $\downarrow$  & $\delta \!>\!1.25$ $\uparrow$ \\      
    \midrule
     DPT &  0.286 & 0.475 & 0.247 & 0.559  & 8.58 & 0.169  \\
     \sname(ours) & \textbf{0.272} & \textbf{0.500} & \textbf{0.244} & \textbf{0.571} & \textbf{7.16} & \textbf{0.190}  \\
    \bottomrule
  \end{tabular}
  \vspace{-10pt}
\end{table*}
\section{Additional Visualizations}
\label{appx:visual}



\subsection{More visualizations of depth maps}


\begin{figure*}[ht!]
    \newcommand{\imageheight}{52pt}
    
    \newcommand{\rotatelabel}[1]{%
      \makebox[0pt][r]{%
        \vbox to \imageheight{%
          \vfil
          \hbox{\rotatebox[origin=c]{90}{\raisebox{0pt}[1ex][1ex]{\small #1}}}
          \vfil
        }%
      }\hspace{-4pt}
    }

  \raggedleft
  \footnotesize
  \setlength{\tabcolsep}{0.5pt}   
  \renewcommand{\arraystretch}{1.0}
  \begin{tabular}{@{} l ccc ccc ccc @{}}
    
       \rotatelabel{Image} &
    \includegraphics[height=\imageheight]{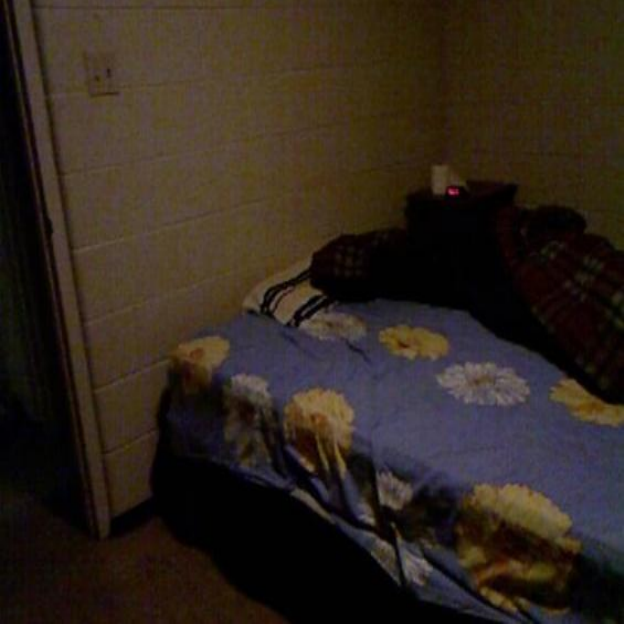} &
    \includegraphics[height=\imageheight]{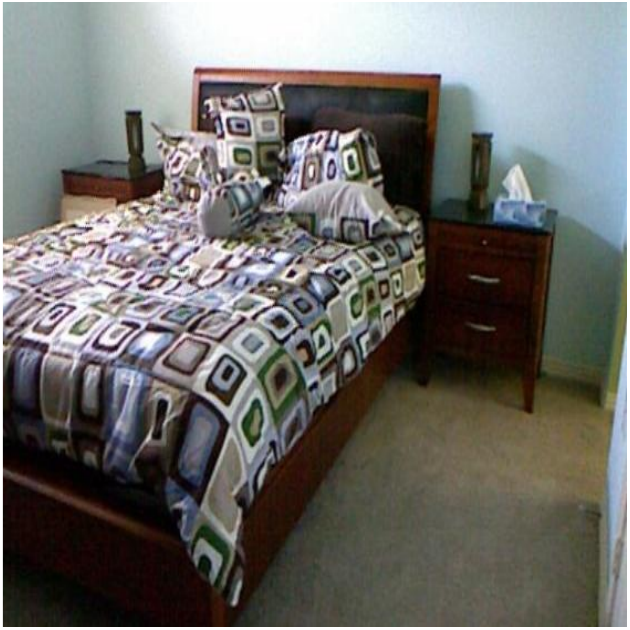} &
    \includegraphics[height=\imageheight]{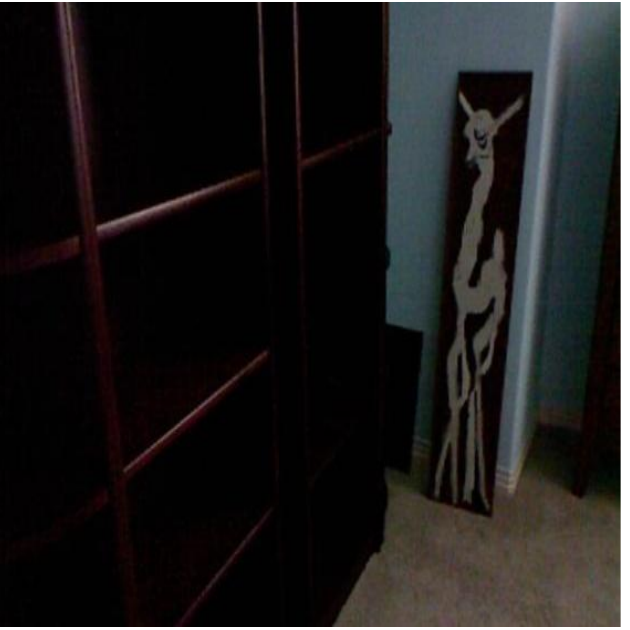} &
        
    \includegraphics[height=\imageheight]{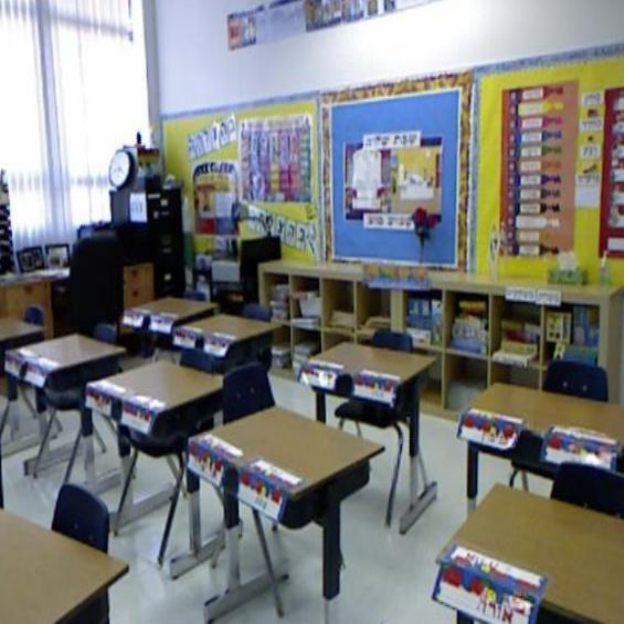} &
    \includegraphics[height=\imageheight]{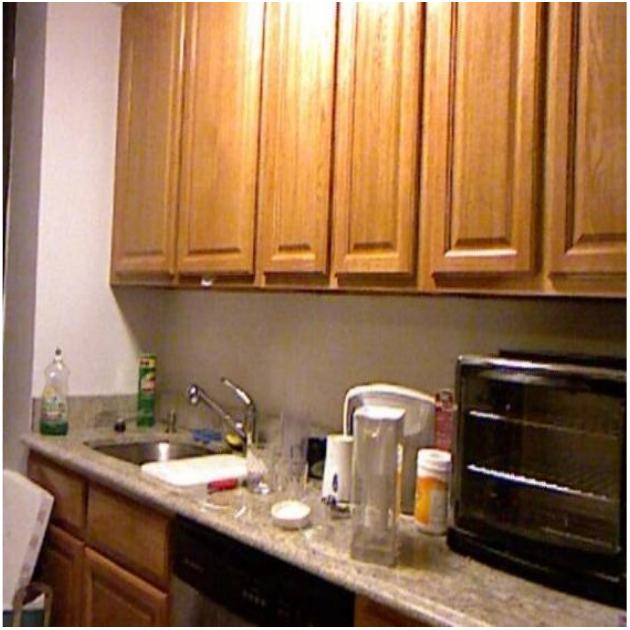} &
    \includegraphics[height=\imageheight]{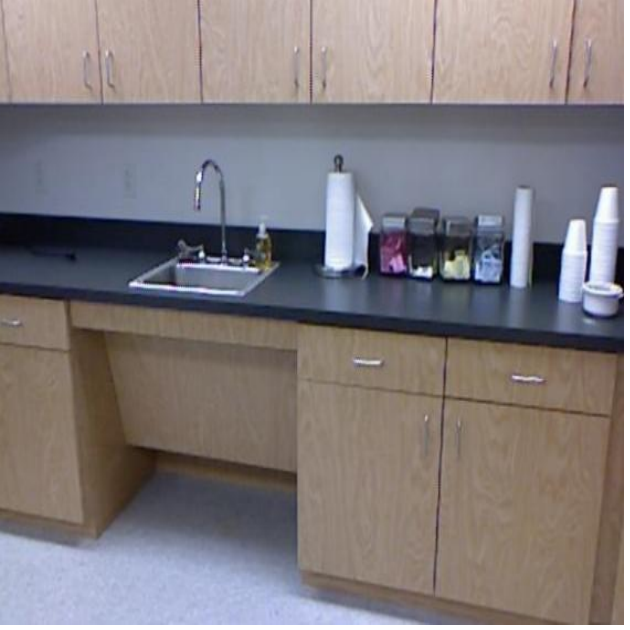} &
    \includegraphics[height=\imageheight]{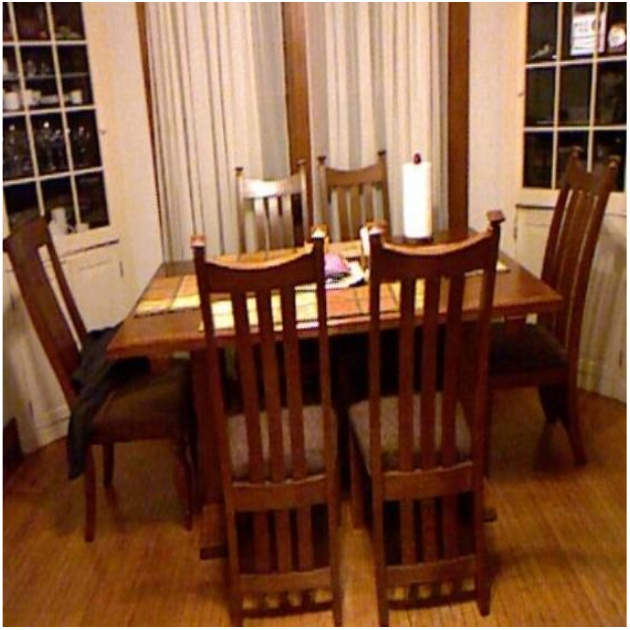} &
    \includegraphics[height=\imageheight]{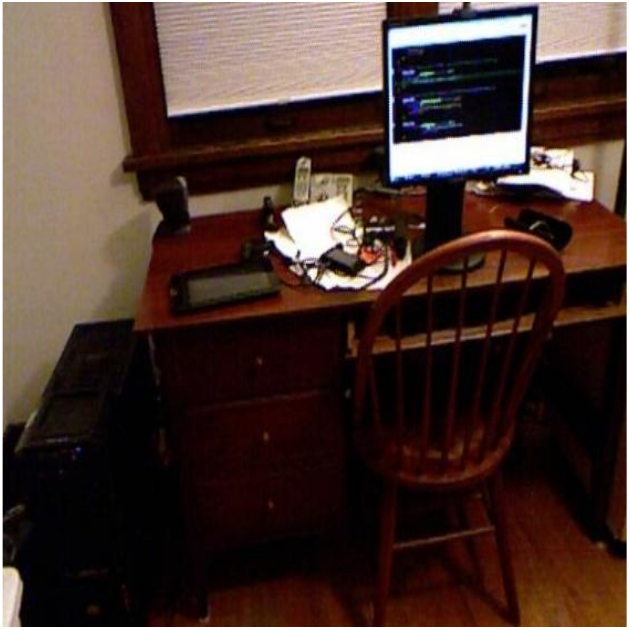} &
    \includegraphics[height=\imageheight]{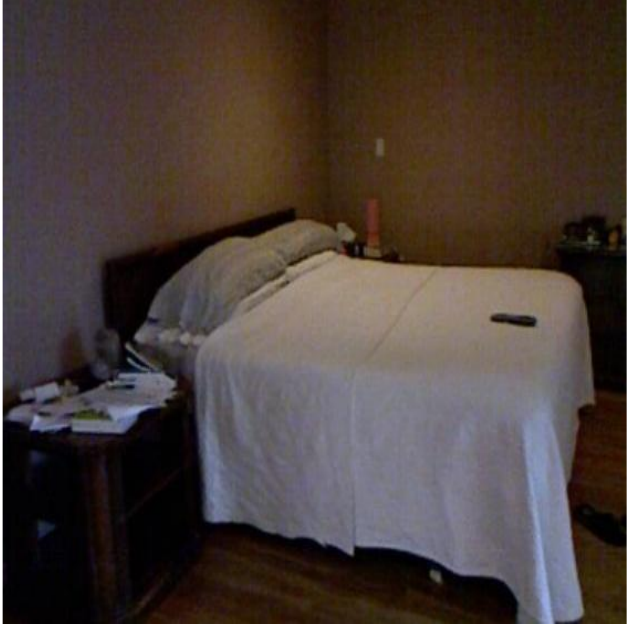} 
    \\

    \noalign{\vskip -2.0pt}
    \rotatelabel{DPT} &
    \includegraphics[height=\imageheight]{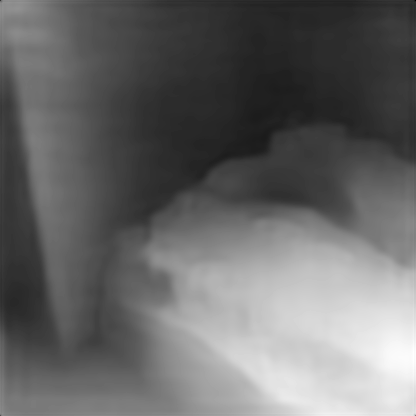} &
    \includegraphics[height=\imageheight]{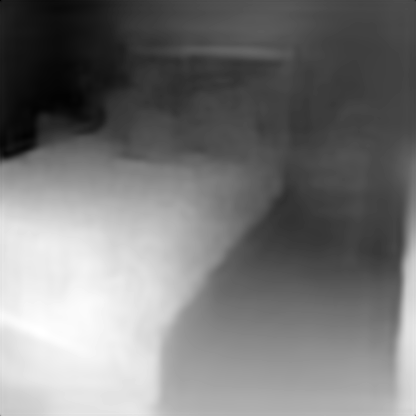} &
    \includegraphics[height=\imageheight]{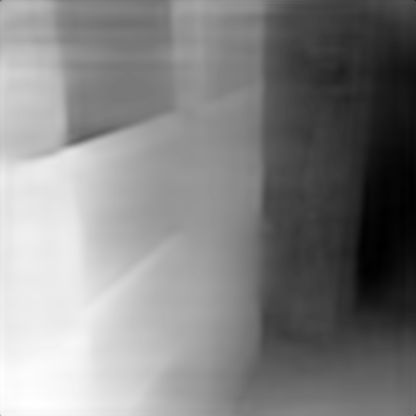} &
    \includegraphics[height=\imageheight]{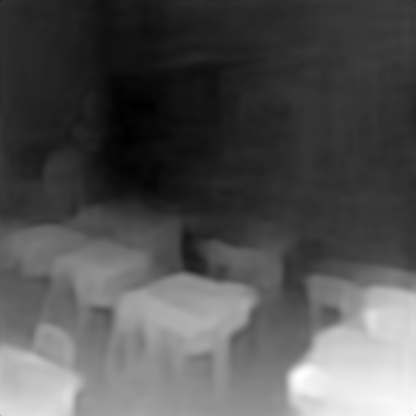} & 
    \includegraphics[height=\imageheight]{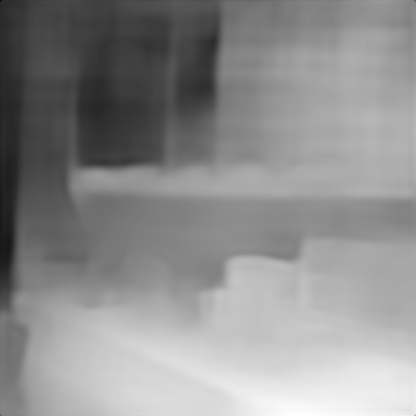} &
     \includegraphics[height=\imageheight]{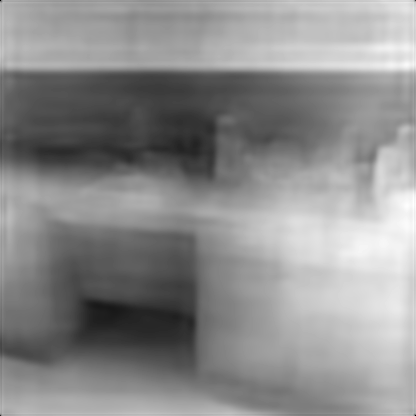} & 
     \includegraphics[height=\imageheight]{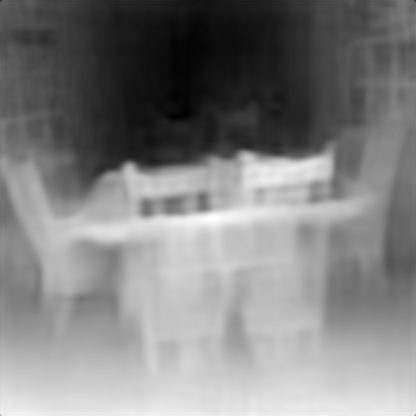} &
     \includegraphics[height=\imageheight]{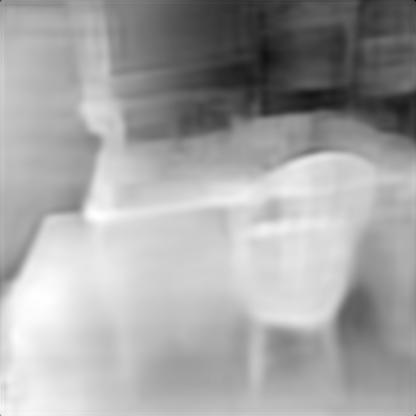} &
      \includegraphics[height=\imageheight]{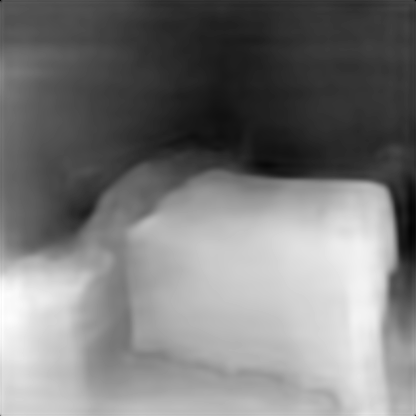}
     \\

    \noalign{\vskip -2.0pt}
    \rotatelabel{\sname (ours)} &
    \includegraphics[height=\imageheight]{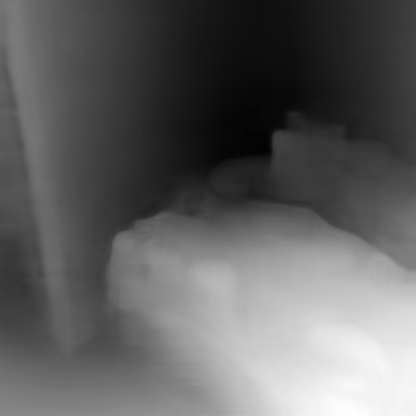} &
    \includegraphics[height=\imageheight]{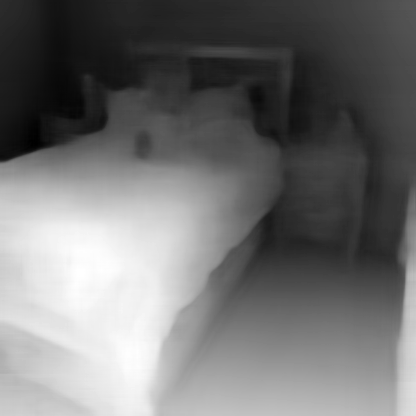} &
    \includegraphics[height=\imageheight]{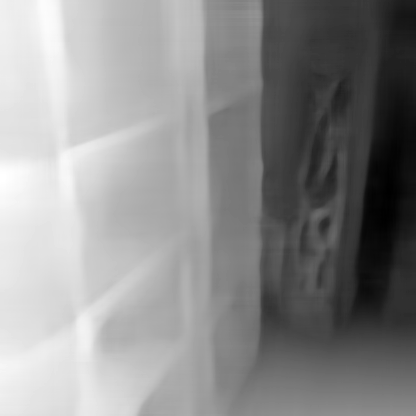} &
    \includegraphics[height=\imageheight]{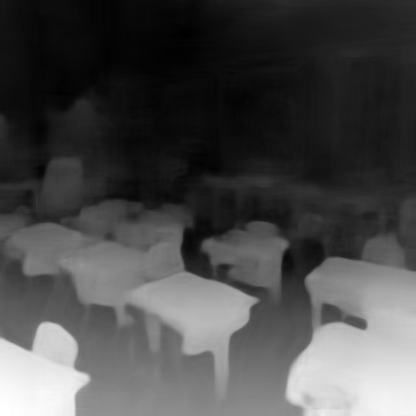} & 
    \includegraphics[height=\imageheight]{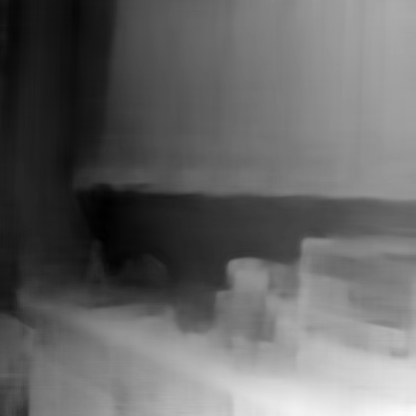} &
     \includegraphics[height=\imageheight]{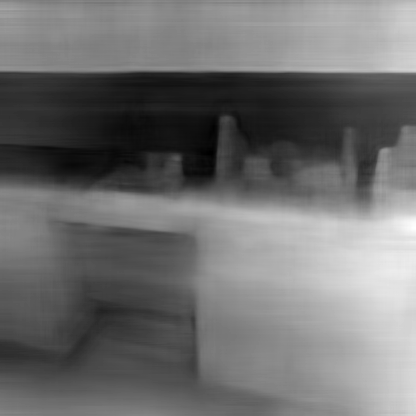} &
     \includegraphics[height=\imageheight]{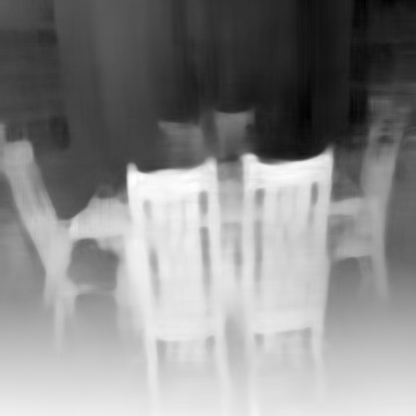} & 
     \includegraphics[height=\imageheight]{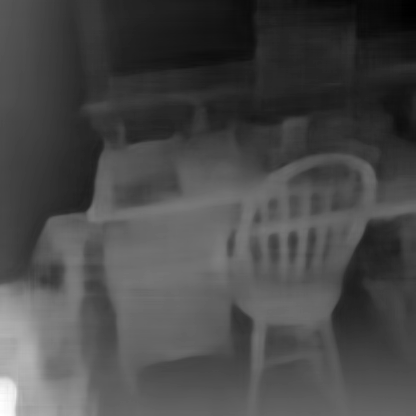} &
     \includegraphics[height=\imageheight]{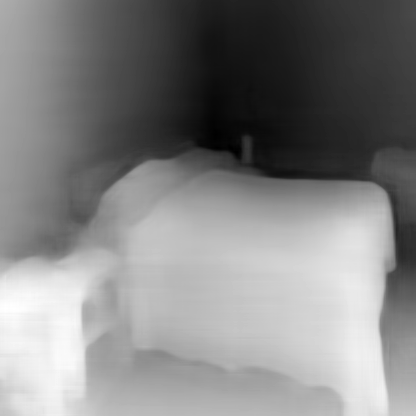} \\

    \noalign{\vskip -2.0pt}
    \rotatelabel{GT} &
    \includegraphics[height=\imageheight]{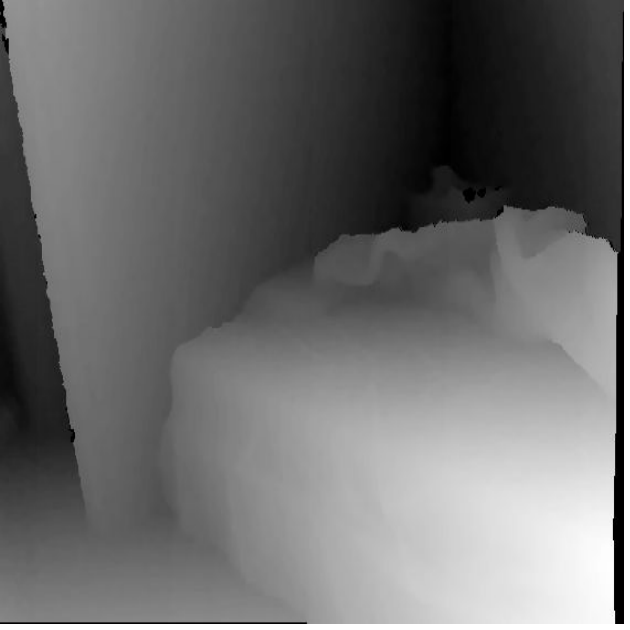} &
    \includegraphics[height=\imageheight]{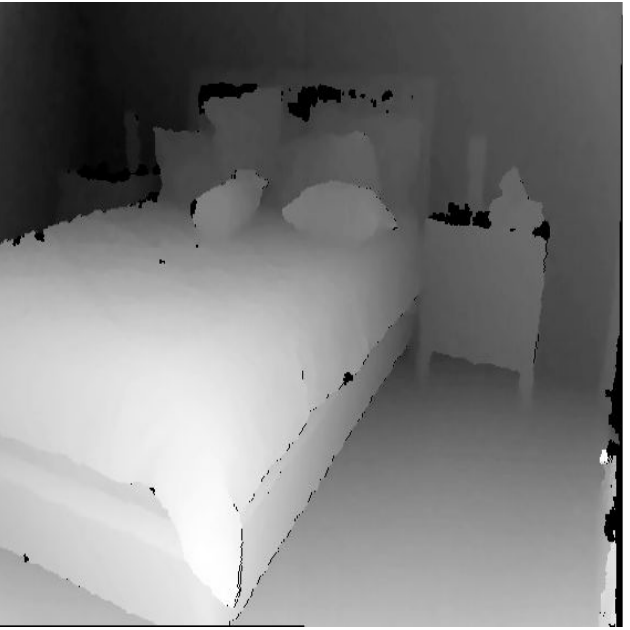} &
    \includegraphics[height=\imageheight]{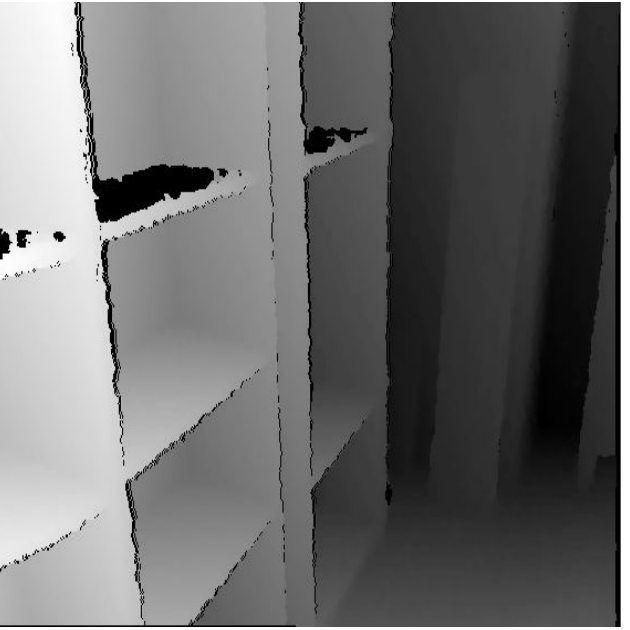} &
        
    \includegraphics[height=\imageheight]{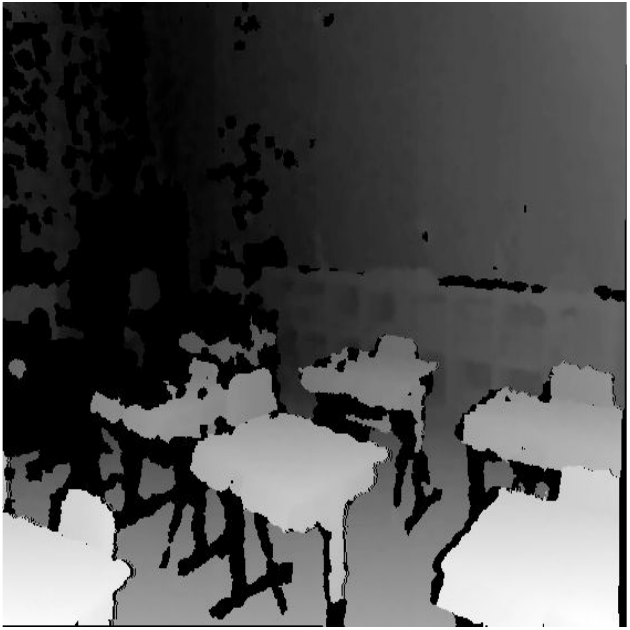} &
    \includegraphics[height=\imageheight]{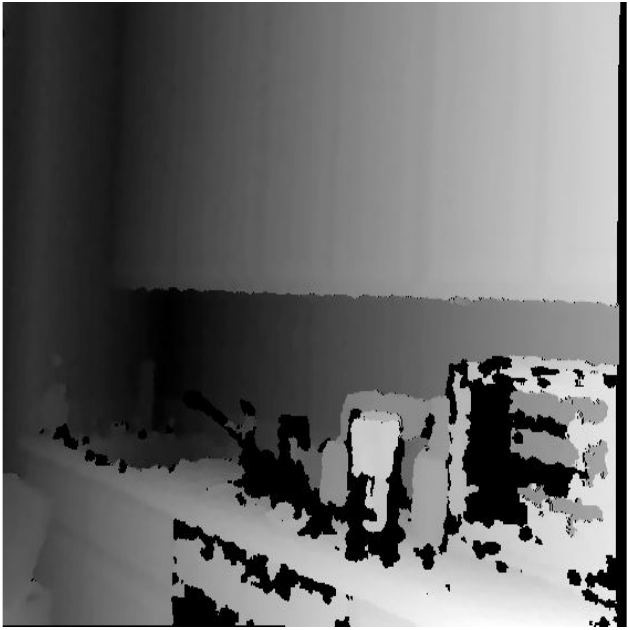} &
    \includegraphics[height=\imageheight]{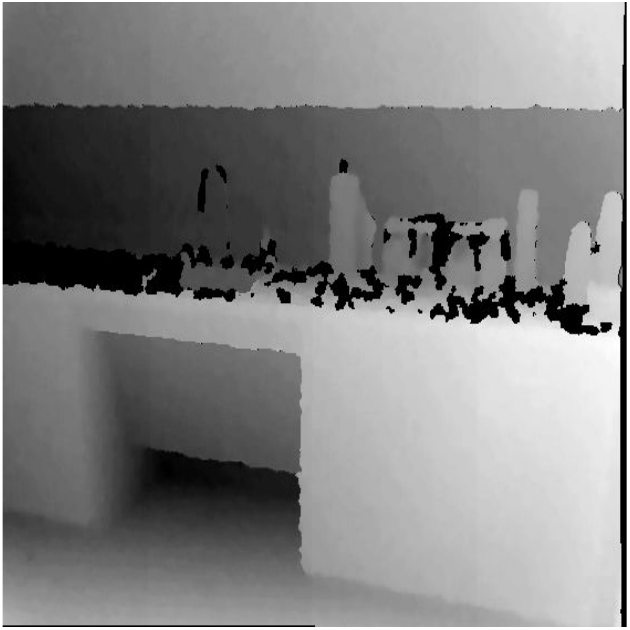} &
    \includegraphics[height=\imageheight]{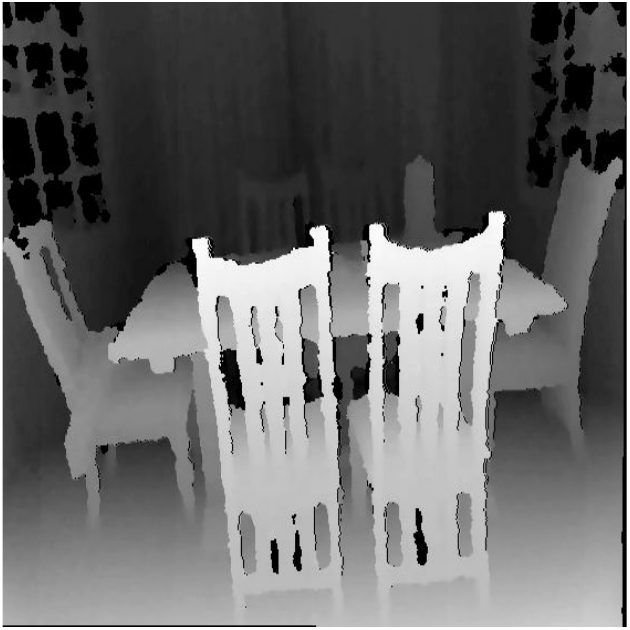} &
    \includegraphics[height=\imageheight]{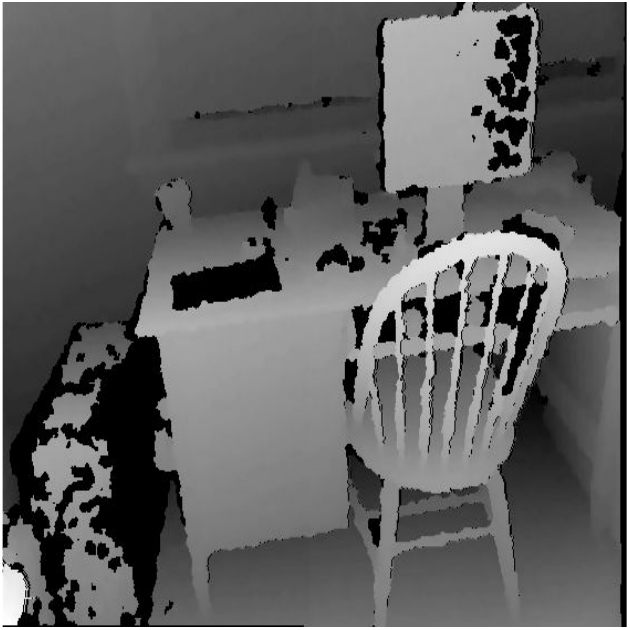} &
    \includegraphics[height=\imageheight]{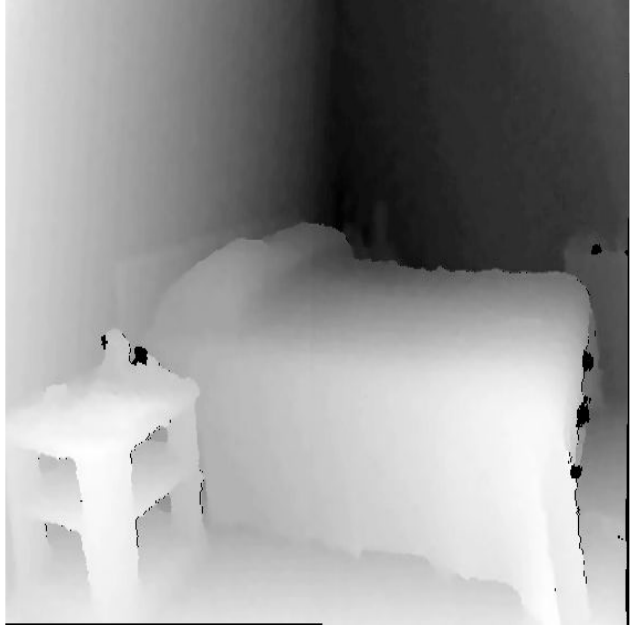}\\

    \end{tabular} 
  \caption{
\textbf{Comparison of depth maps.} \sname captures more accurate shapes with sharper boundaries, whereas DPT produces blurrier results. 
}\label{fig:fig:more-depth}
\end{figure*}

\clearpage

\subsection{More visualizations of image retrieval}

\begin{figure*}[ht!]
    \newcommand{\myheight}{50pt}
    
    \newcommand{\mylabel}[1]{%
      \makebox[0pt][r]{%
        \vbox to \myheight{%
          \vfil
          \hbox{\rotatebox[origin=c]{90}{\raisebox{0pt}[1ex][1ex]{\small #1}}}
          \vfil
        }%
      }\hspace{-4pt}
    }
    \newcommand{\myimage}[1]{\includegraphics[height=\myheight]{fig/retrieval/#1.jpg}}

  \raggedleft
  \footnotesize
  \setlength{\tabcolsep}{0.5pt}   
  \renewcommand{\arraystretch}{1.0}
  \begin{tabular}{@{} l ccc ccc ccc @{}}
    
    \mylabel{Query} &
    \myimage{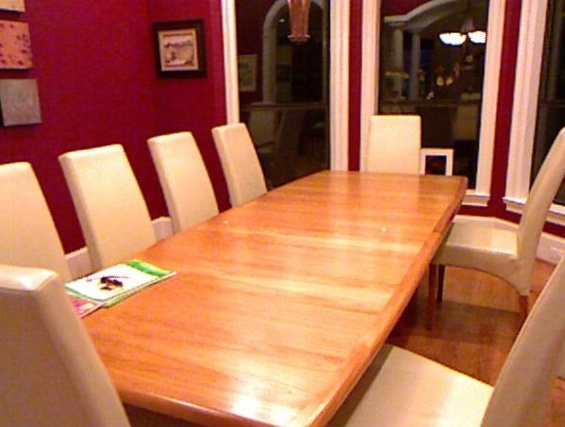} &
    \myimage{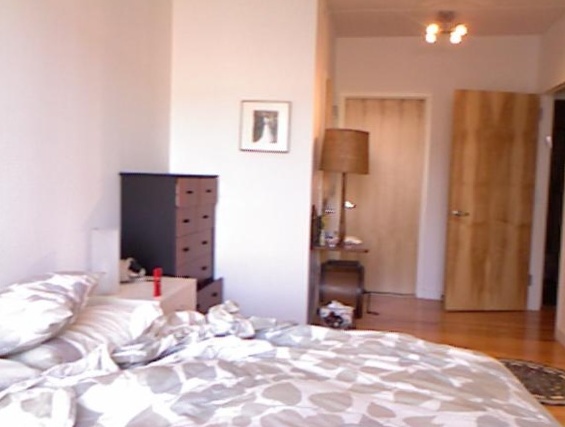} &
    \myimage{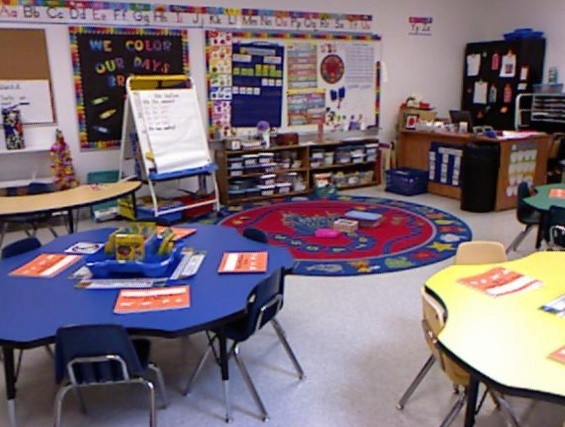} &
    \myimage{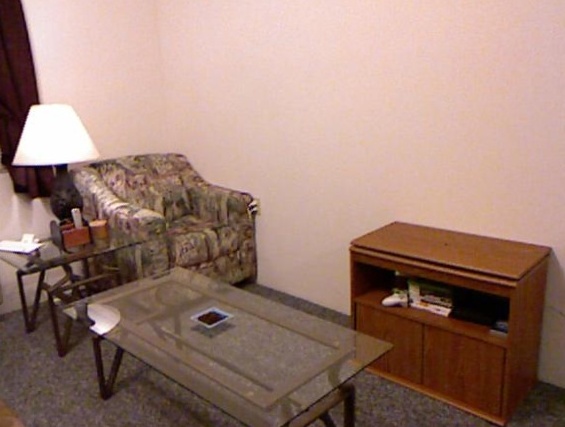} &
    \myimage{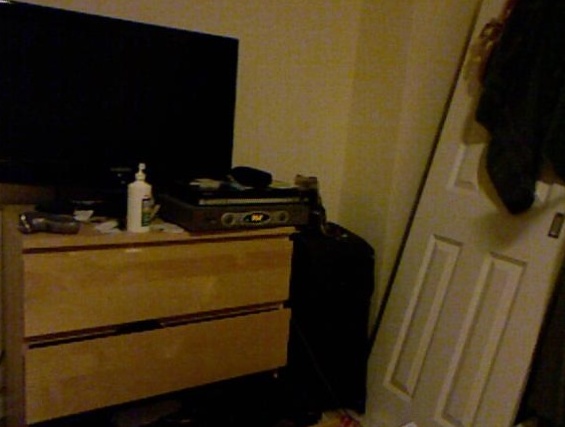} &
    \myimage{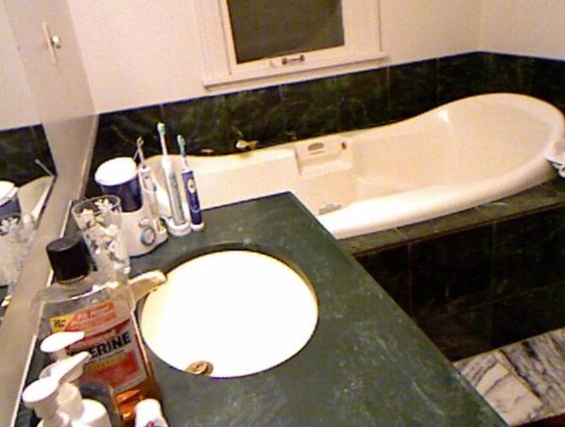} &
    \myimage{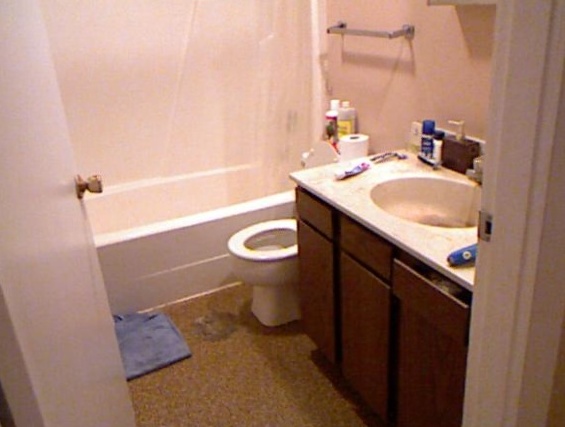}
    \\

    \noalign{\vskip -2.0pt}
    \mylabel{\sname (ours)} &
    \myimage{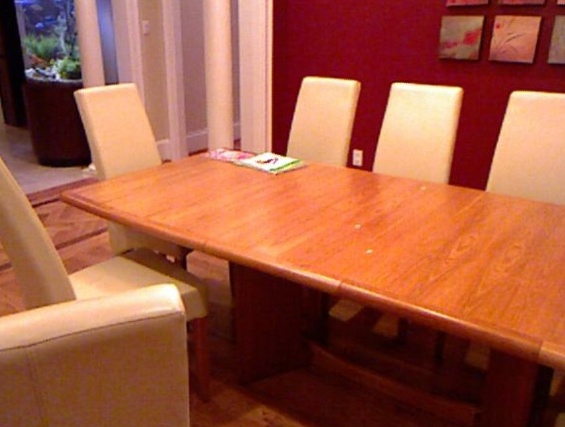} &
    \myimage{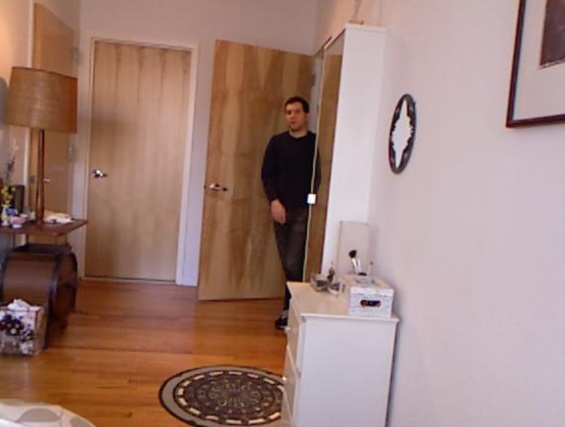} &
    \myimage{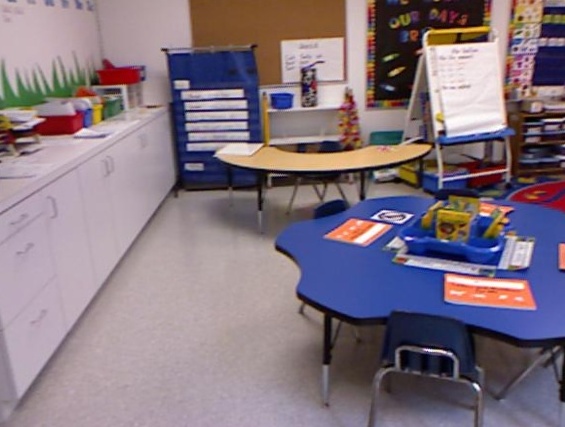} &
    \myimage{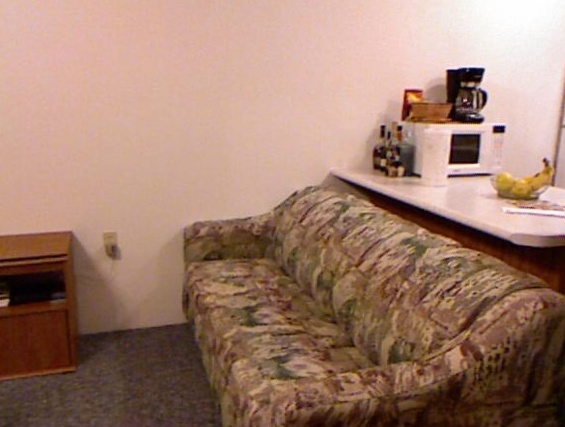} &
    \myimage{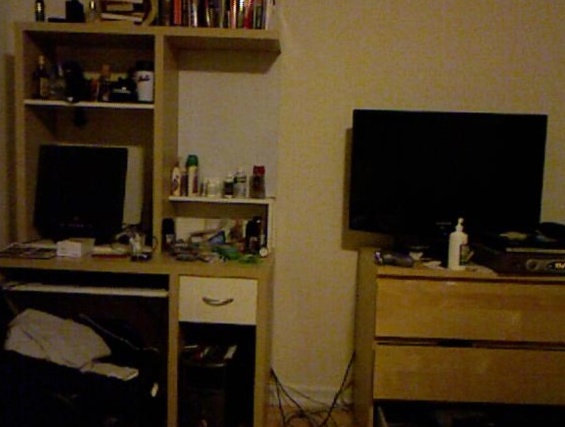} &
    \myimage{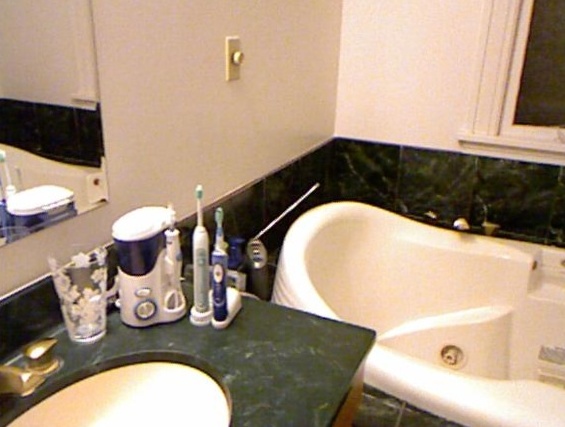} &
    \myimage{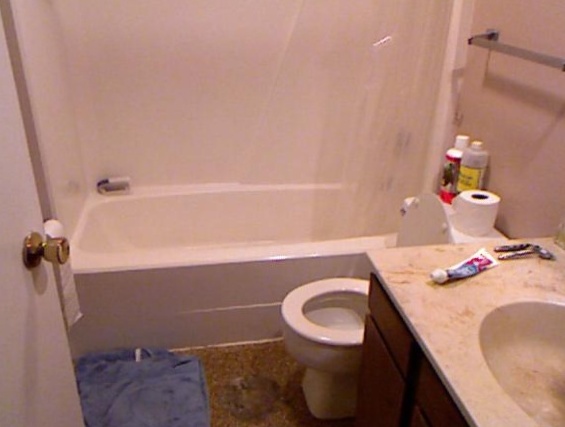}
    \\

    \noalign{\vskip -2.0pt}
    \mylabel{DPT} &
    \myimage{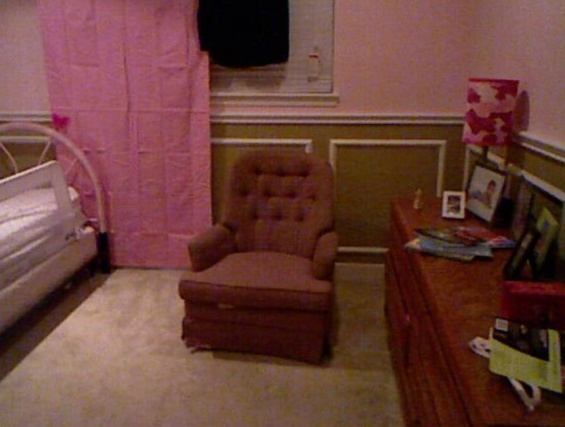} &
    \myimage{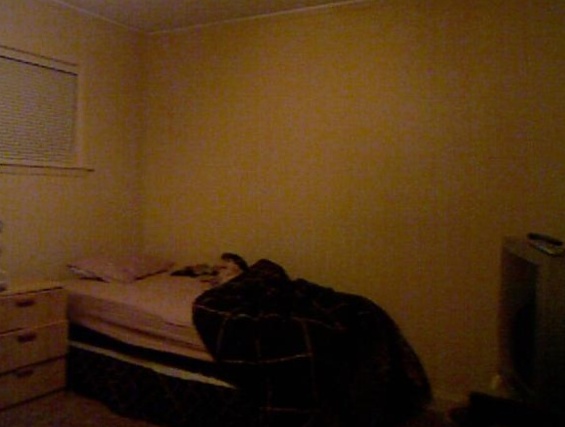} &
    \myimage{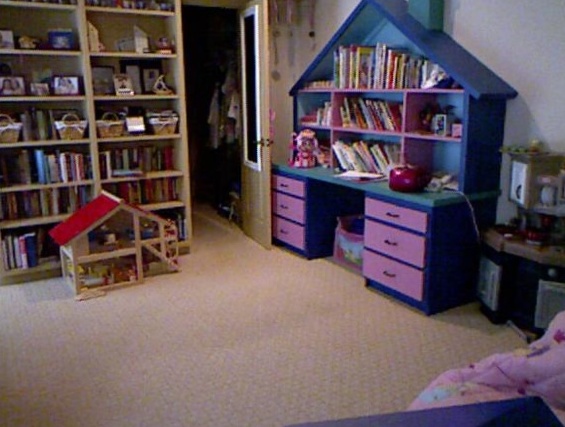} &
    \myimage{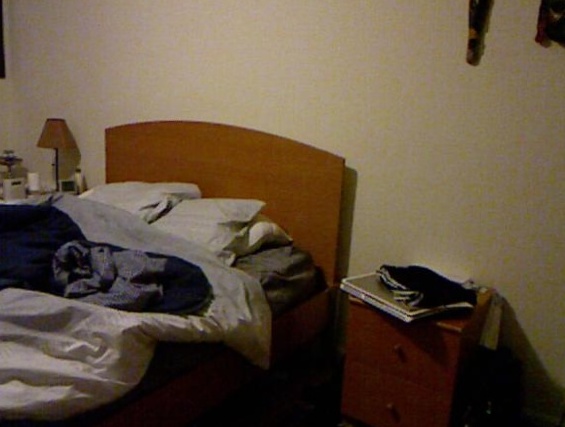} &
    \myimage{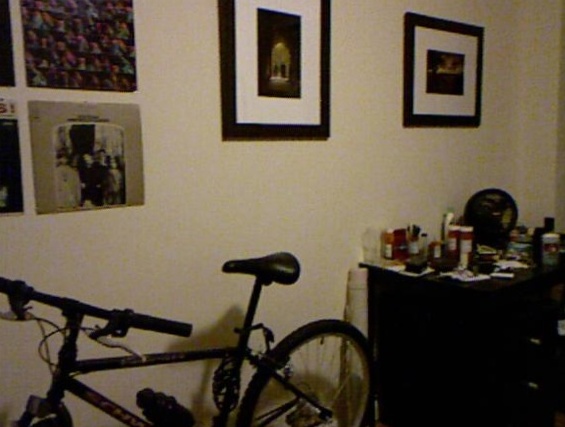} &
    \myimage{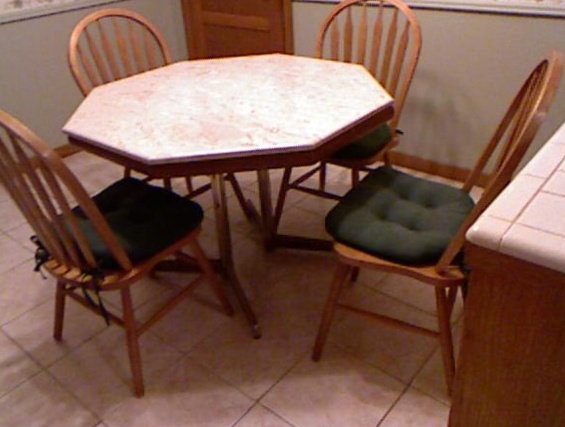} &
    \myimage{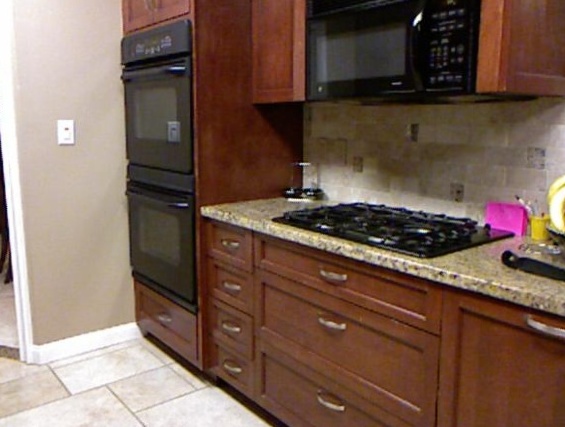}
     \\
    
    \end{tabular} 
  \caption{
  \textbf{Comparison of top-1 image retrieval results.} \sname retrieves samples that are more structurally similar to the query, indicating that its global embedding effectively captures scene layouts. In contrast, DPT focuses more on visual appearance, as shown in column 3, where it retrieves an image from a different scene that shares a similar color of blue.
}\label{fig:fig:more-retrieval}
\end{figure*}

\subsection{More visualizations of 3D reconstruction}

\begin{figure*}[ht!]
  \centering\footnotesize
  \setlength{\tabcolsep}{10pt}   
  \renewcommand{\arraystretch}{1.0}

    \newcommand{\overlayimage}[3]{%
    \tikz[baseline]{
      \node[inner sep=0pt] (img) {\includegraphics[trim=5 12.4 5 12.4, clip, width=#1]{#2}};
      \node[anchor=south east, font=\footnotesize, fill=white, text opacity=0.8, draw=black, rounded corners=1pt, xshift=-2pt, yshift=2pt] 
        at (img.south east) {#3};
    }}

  \newcommand{\hspacesmall}{\hspace{1.5pt}}
  \newcommand{\hspacelarge}{\hspace{2.5pt}}
  \newcommand{\hspacelarger}{\hspace{5pt}}
 
  \begin{tabular}{
  @{}
  c@{\hspacelarge}
  c@{\hspacesmall}c@{\hspacesmall}c@{\hspacelarger}
  c@{\hspacelarger}c@{\hspacelarger}c@{}
  }
    Image &
    DPT & \sname (ours) & GT &
    DPT & \sname (ours) & GT \\[1pt]

    \includegraphics[width=0.13\textwidth]{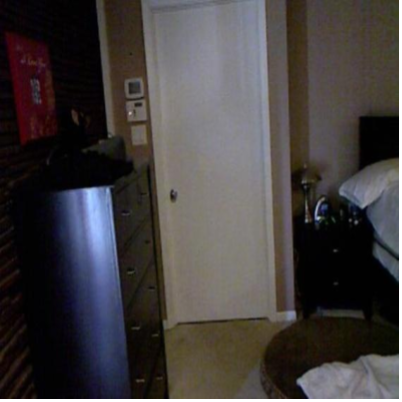} &
    \includegraphics[trim=5 12.4 5 12.4, clip, width=0.135\textwidth]{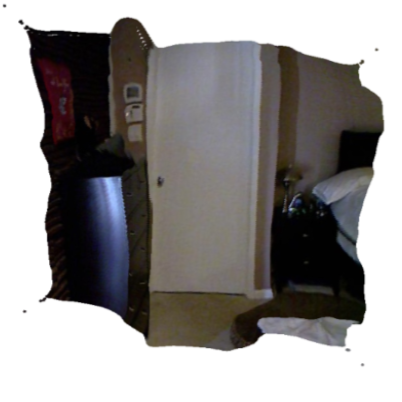} &
    \includegraphics[trim=5 12.4 5 12.4, clip, width=0.135\textwidth]{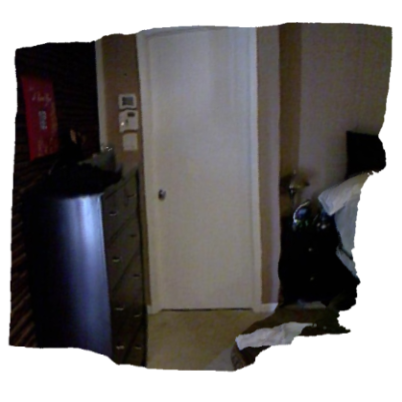} &
    \includegraphics[trim=5 12.4 5 12.4, clip, width=0.135\textwidth]{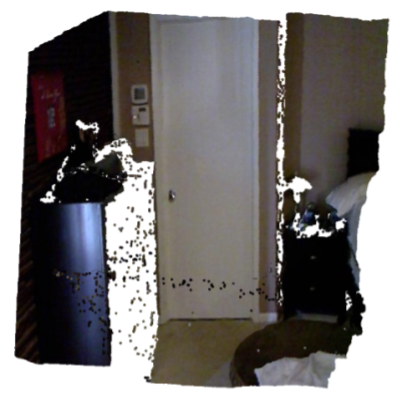} & 
    \includegraphics[width=0.135\textwidth]{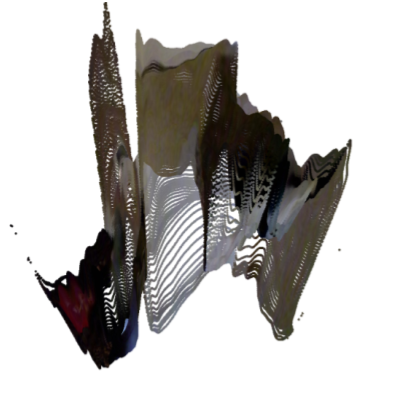} &
    \includegraphics[trim=5 12.4 5 12.4, clip, width=0.135\textwidth]{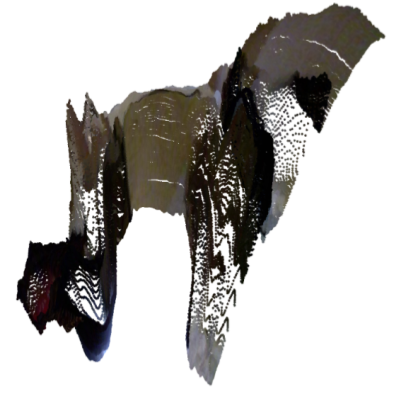} &
    \includegraphics[width=0.135\textwidth]{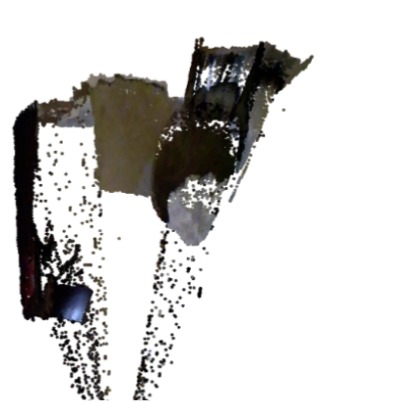}
     \\
     \includegraphics[width=0.13\textwidth]{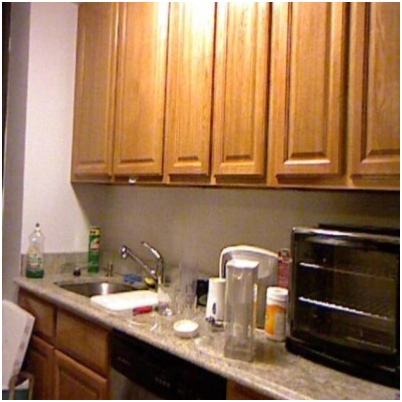} &
    \includegraphics[trim=5 12.4 5 12.4, clip, width=0.135\textwidth]{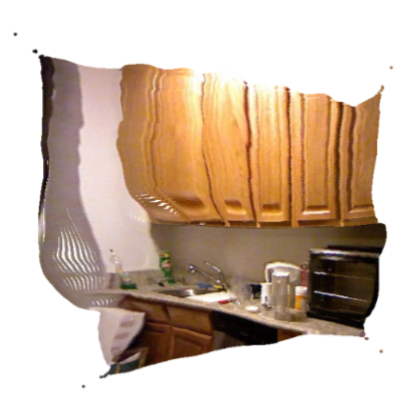} &
    \includegraphics[trim=5 12.4 5 12.4, clip, width=0.135\textwidth]{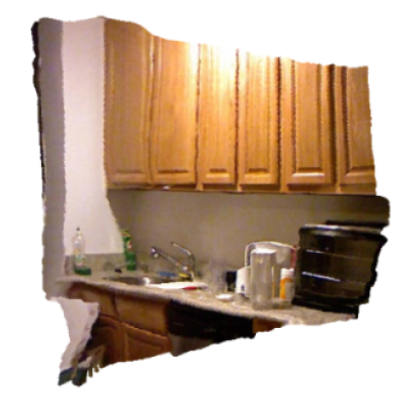} &
    \includegraphics[trim=5 12.4 5 12.4, clip, width=0.135\textwidth]{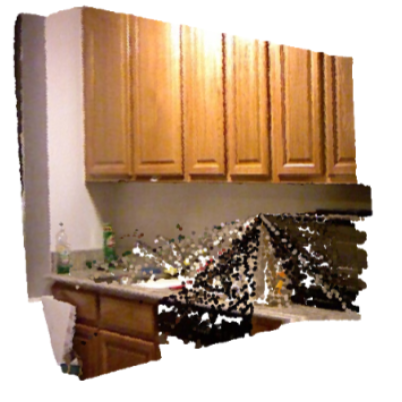} & 
    \includegraphics[width=0.135\textwidth]{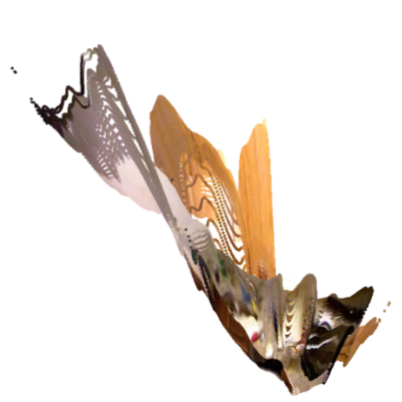} &
    \includegraphics[trim=5 12.4 5 12.4, clip, width=0.135\textwidth]{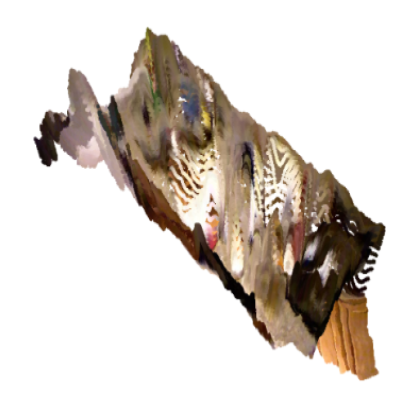} &
    \includegraphics[width=0.135\textwidth]{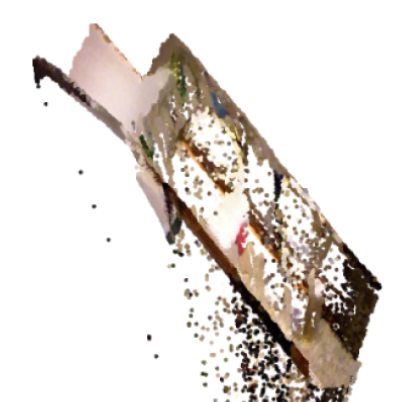}
     \\
     \includegraphics[width=0.13\textwidth]{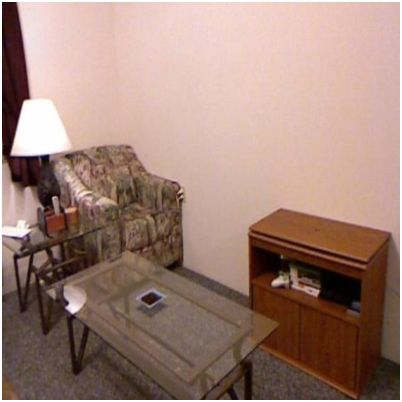} &
    \includegraphics[trim=5 12.4 5 12.4, clip, width=0.135\textwidth]{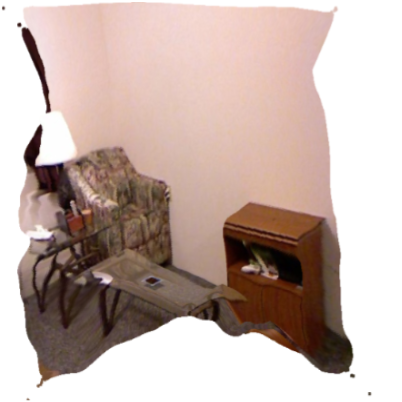} &
    \includegraphics[trim=5 12.4 5 12.4, clip, width=0.135\textwidth]{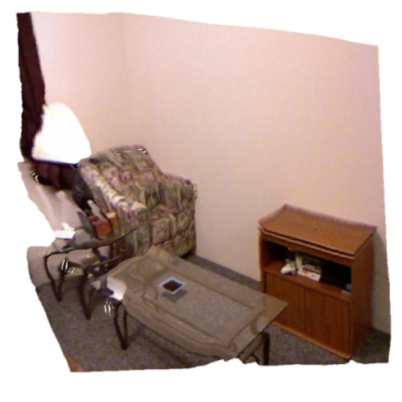} &
    \includegraphics[trim=5 12.4 5 12.4, clip, width=0.135\textwidth]{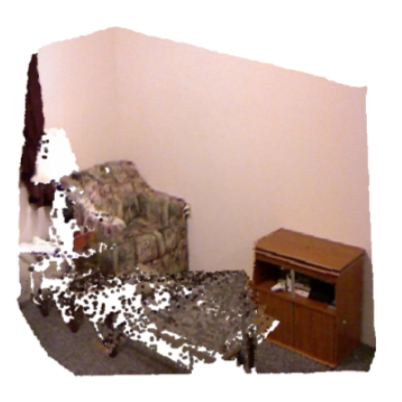} & 
    \includegraphics[width=0.135\textwidth]{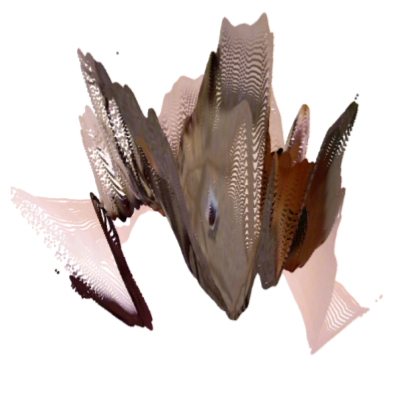} &
    \includegraphics[trim=5 12.4 5 12.4, clip, width=0.135\textwidth]{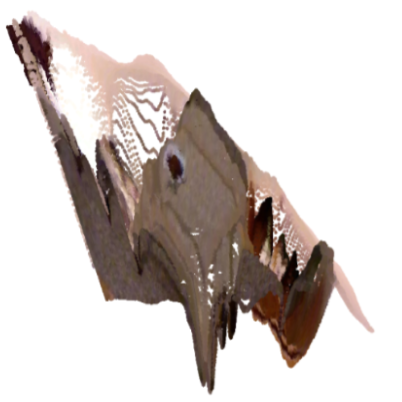} &
    \includegraphics[width=0.135\textwidth]{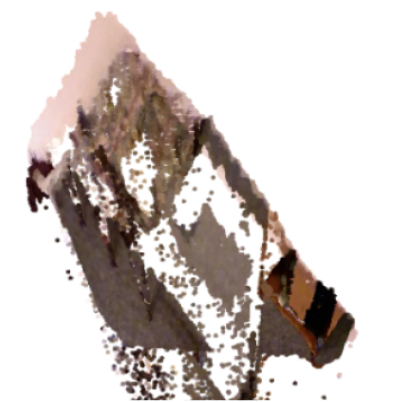}
     \\
     \includegraphics[width=0.13\textwidth]{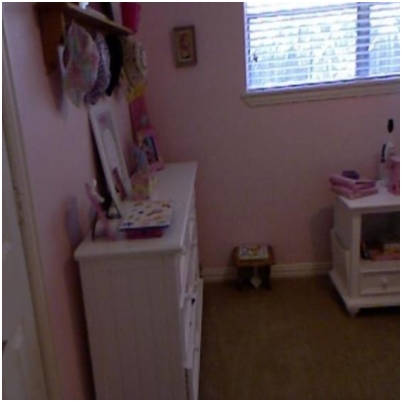} &
    \includegraphics[trim=5 12.4 5 12.4, clip, width=0.135\textwidth]{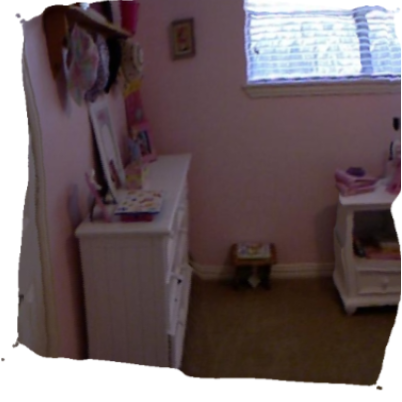} &
    \includegraphics[trim=5 12.4 5 12.4, clip, width=0.135\textwidth]{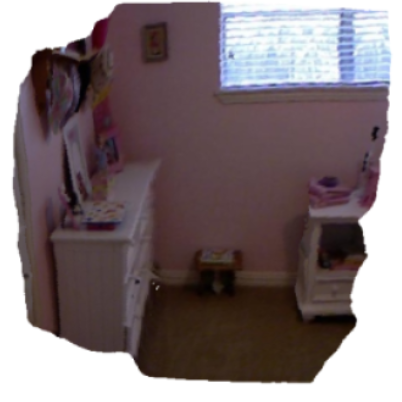} &
    \includegraphics[trim=5 12.4 5 12.4, clip, width=0.135\textwidth]{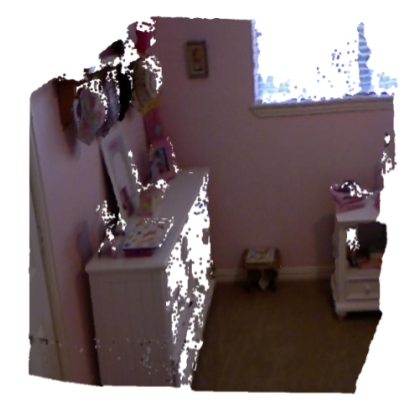} & 
    \includegraphics[width=0.135\textwidth]{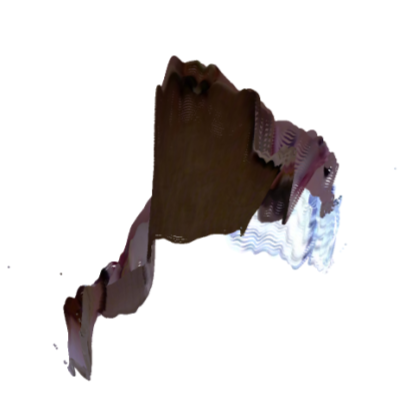} &
    \includegraphics[trim=5 12.4 5 12.4, clip, width=0.135\textwidth]{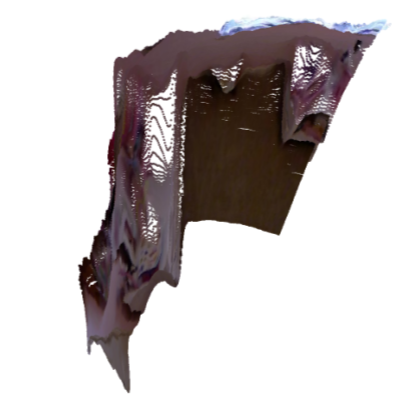} &
    \includegraphics[width=0.135\textwidth]{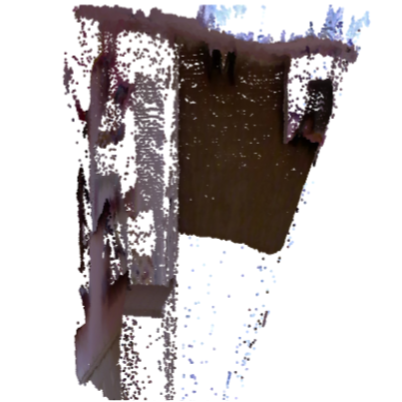}
     \\
    
  \end{tabular}
    \caption{%
\textbf{Comparison of 3D reconstruction results.}
Frontal views (cols 2–4) and bird’s-eye views (cols 5–7). DPT yields curved wall boundaries in the frontal views, which lead to distorted 3D reconstructions visible in the bird’s-eye views. In contrast, \sname produces sharp depth edges that preserve straight object contours and more faithfully represent the ground-truth 3D geometry.
}

    \label{fig:fig:point}

\vspace{-2pt}

\end{figure*}

\section{Limitations and Broader Impacts}
\label{appx:limitation}


\subsection{Limitations and future works}
While \sname demonstrates strong structural coherence in depth estimation, its hierarchical design requires the number of tokens to be pooled at each level to be specified in advance. As a result, the unpooling process assumes a fixed correspondence between pooled and unpooled tokens, which may limit flexibility when the underlying scene structure does not align well with the predefined hierarchy. This design choice constrains the model to operate within a fixed hierarchical schedule, potentially reducing adaptability to scenes with highly varying structural complexity. A promising direction for future work is to relax this constraint by enabling adaptive or data-driven determination of the hierarchical schedule, allowing the model to adjust the number of tokens dynamically based on scene complexity. In addition, exploring more flexible unpooling mechanisms that do not require a strict one-to-one correspondence between pooled and unpooled tokens could further improve robustness to diverse scene layouts.





\subsection{Broader impacts}

Structured understanding of the 3D world and accurate depth estimation are central challenges in AI, with direct impact on safety-critical applications such as autonomous driving, augmented reality, and robotics. In practice, failures in these systems often result not from a lack of data but from insufficient structured reasoning, making predictions vulnerable to occlusion, unusual viewpoints, and dynamic environments. Our framework promotes geometry-aware perception by producing robust and interpretable depth estimates that align with scene structure. This can improve reliability in complex real-world settings and lead to systems with more transparent failure modes.

\end{document}